\documentclass{article}
\usepackage{ijcai24}

\usepackage{times}
\usepackage{soul}
\usepackage{url}
\usepackage[hidelinks]{hyperref}
\usepackage[utf8]{inputenc}
\usepackage[small]{caption}
\usepackage{graphicx}
\usepackage{amsmath}
\usepackage{amsthm}
\usepackage{booktabs}
\usepackage{algorithm}
\usepackage{algorithmic}
\usepackage[switch]{lineno}
\usepackage{color}
\usepackage[table]{xcolor}
\usepackage{multirow}
\usepackage{threeparttable}

\usepackage{amssymb}%for \checkmark
\title{RoboFusion: Towards Robust Multi-Modal 3D Object Detection via SAM}

\author{
Ziying Song$^1$
\and
Guoxing Zhang$^2$\and
Lin Liu$^1$\and
Lei Yang$^3$\and
Shaoqing Xu$^4$\and
Caiyan Jia$^1$\thanks{Corresponding author}\and \\
Feiyang Jia$^1$\and
Li Wang$^5$
\affiliations
\small $^1$School of Computer Science and Technology \& Beijing Key Lab of Traffic Data Analysis and Mining, Beijing Jiaotong University\\
$^2$Hebei University of Science and Technology
$^3$Tsinghua University
$^4$University of Macau
$^5$Beijing Institute of Technology
\emails
\tt \small\{songziying, cyjia, feiyangjia\}@bjtu.edu.cn
}
\begin{document}

\maketitle

\begin{abstract}
 Multi-modal 3D object detectors are dedicated to exploring secure and reliable perception systems for autonomous driving (AD). %However, %while 
 Although achieving state-of-the-art (SOTA) performance on clean benchmark datasets, they tend to overlook the complexity and harsh conditions of real-world environments. %Meanwhile, 
 With the emergence of visual foundation models (VFMs), opportunities and challenges are presented for improving the robustness and generalization of multi-modal 3D object detection in AD. %autonomous driving. 
 Therefore, we propose \textbf{RoboFusion}, a robust framework that leverages VFMs like SAM to tackle out-of-distribution (OOD) noise scenarios. We first adapt the original SAM for AD %autonomous driving 
 scenarios named \textbf{SAM-AD}. To align SAM or \textbf{SAM-AD} with multi-modal methods, we then introduce \textbf{AD-FPN} for upsampling the image features extracted by SAM. We employ wavelet decomposition to denoise the depth-guided images for further noise reduction and weather interference. At last, we employ self-attention mechanisms to adaptively reweight the fused features, enhancing informative features while suppressing excess noise. In summary, %our 
 RoboFusion %gradually 
 significantly reduces noise by leveraging the generalization and robustness of VFMs, thereby enhancing the resilience of multi-modal 3D object detection. Consequently, %our 
 RoboFusion achieves %
%state-of-the-art 
 SOTA performance in noisy scenarios, as demonstrated by the KITTI-C and nuScenes-C benchmarks. Code is available at \url{https://github.com/adept-thu/RoboFusion}.
\end{abstract}

\section{Introduction}

\begin{figure*}[t]
\centering
\includegraphics[width=\textwidth]{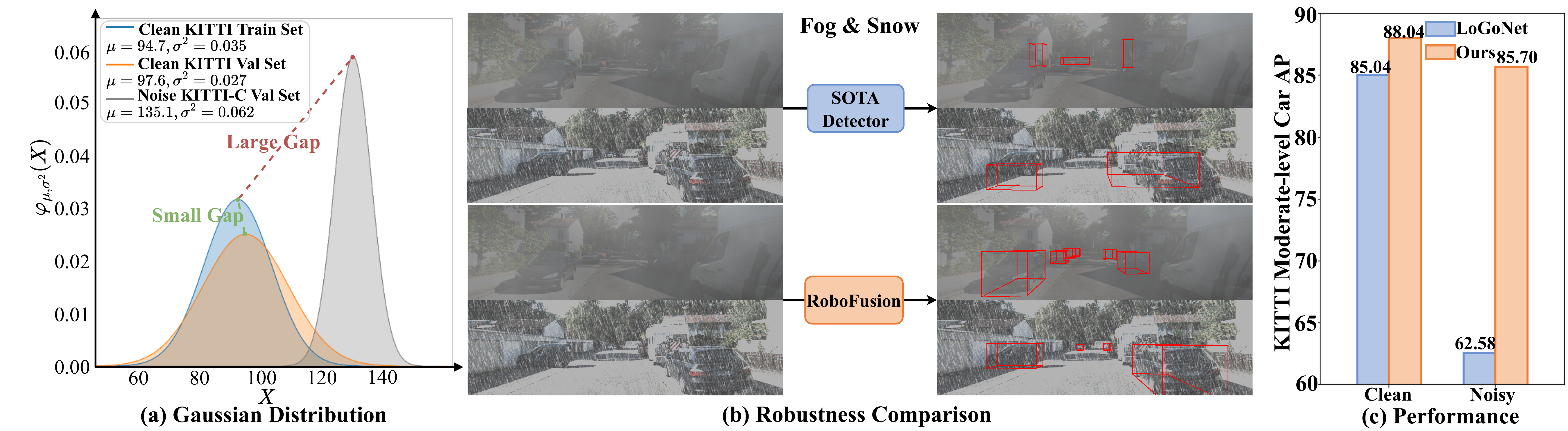}

\caption[ ]{\textbf{(a)} We employ Gaussian distributions to represent the distributional disparities among the datasets. Indeed, there exists a large gap in data distribution between an OOD noise validation set and a clean validation set. Where the X-axis represents the set of mean pixel values in a dataset, \(X = \{x_{i} \,|\, i=1,2,...,N\}\), with \(x_{i} = \frac{1}{H \times W \times 3} \sum_{i=1}^{H} \sum_{j=1}^{W} \sum_{k=1}^{3}(I_{ijk})\), where \(N\) is the number of the dataset, \(H\) is the height, \(W\) is the width, and \(I_{ijk}\) denotes the pixel values for each image. \textbf{(b)} Visual foundation models (VFMs) like SAM \cite{sam}, show robust performance in many noisy scenarios. Yet, the current methods are not robust enough to %robustly 
predict 3D tasks for autonomous driving perception. \textbf{(c)} To this end, we propose a robust framework, RoboFusion, which employs VFMs at the SOTA multi-modal 3D object detection. Empirical results reveal that our method surpasses the Top-performing LoGoNet\cite{logonet} on the KITTI Leaderboard by a margin of 23.12\% mAP (Weather) on KITTI-C \cite{Robustness3d} noisy scenarios. Notably, our RoboFusion shows better performance with LoGoNet \cite{logonet} in clean KITTI \cite{kitti} dataset.}

\label{fig:motivation}
\end{figure*}

Multi-modal 3D object detection plays a pivotal role in autonomous driving (AD) \cite{wang2023multi,song2024robustness}. Different modalities often provide complementary information. For instance, images contain richer semantic representations, yet lack depth information. In contrast, point clouds offer geometric and depth details, but they are sparse and lack semantic information.
Therefore, effectively leveraging the advantages of multi-model while mitigating their limitations contributes to enhancing the robustness and accuracy of perception systems \cite{song2023graphalign}.

With the emergence of AD datasets \cite{kitti,nuscenes,zhang2023dual}, state-of-the-art (SOTA) methods \cite{bevfusion-mit,transfusion,focalconv,epnet,logonet,song2024voxelnextfusion} on `clean' datasets \cite{kitti,nuscenes} have achieved record-breaking performance. However, they overlook the exploration of robustness and generalization in out-of-distribution (OOD) scenarios \cite{Robustness3d}. For example, the KITTI dataset \cite{kitti} lacks severe weather conditions. When SOTA methods \cite{focalconv,logonet,bevfusion-mit} learn from these sunny weather datasets, can they truly generalize and maintain robustness in severe weather conditions like snow and fog? 

The answer is `No', as shown in Fig.~\ref{fig:motivation} and verified in Table \ref{tab_KITTI_val_C}. %However, 
People often utilize domain adaptation (DA) techniques to address these challenges \cite{wang2023ssda3d,tsai2023viewer,peng2023cl3d,hu2023density}. Although DA techniques improve the robustness of 3D object detection and reduce the need for annotated data, they have some profound drawbacks, including domain shift limitations, label shift issues, and overfitting risks \cite{oza2023unsupervised}. For instance, DA techniques may be constrained if the differences between two domains are  significant, leading to performance degradation on the target domain.

Recently, both Natural Language Processing (NLP) and Computer Vision (CV) have witnessed the appearance and the power of a series of foundation models \cite{sam,gpt4,fastsam,mobilesam}, resulting in the emergence of new paradigms in deep learning. For example, 
% Kirillov {\it et al.}~ 
a series of novel visual foundation models (VFMs) \cite{sam,fastsam,mobilesam} %in CV %called SAM 
have been developed. Thanks to their extensive training on huge datasets, %, SA-1B \cite{sam}, 
these models exhibit powerful generalization capabilities. These developments have inspired new ideas, leveraging the robustness and generalization abilities of VFMs to achieve generalization in OOD noisy scenarios, much like how adults generalize knowledge when encountering new situations, without relying on DA techniques \cite{wang2023ssda3d,tsai2023viewer}.
% However, some attempts \cite{sam3d} to combine SAM with pre-trained 3D models did not yield impressive results. This is primarily because SAM was originally trained on 2D images and is more suitable for generalization on image-related tasks.

%Although SOTA multi-modal methods \cite{virconv,bevfusion-mit} have achieved high performance on `clean' datasets, they still face challenges in dealing with OOD noise, which hinders their practical applicability in real-world scenarios. 
%Specifically, 
Inspired by the success of VFMs in CV tasks, in this work, we intend to use these models to tackle the challenges of multi-modal 3D object detectors in OOD noise scenarios.
%Our core idea is to achieve robust fusion features by combining more resilient image and point cloud features %obtained through 
%via SAM. %by taking its advantage of zero-shot generalization ability.
Therefore, we propose a robust framework, RoboFusion, which leverages VFMs like SAM to adapt a 3D multi-modal object detector from clean scenarios to OOD noise scenarios.
In particular, the adaptation strategies for SAM are as follows. 1) We utilize features extracted from SAM rather than inference segmentation results. 2) We propose \textbf{SAM-AD}, which is a pre-trained SAM for AD scenarios. 3)  We introduce a novel \textbf{AD-FPN} to address the issue of feature upsampling for aligning VFMs with multi-modal 3D object detector.
% SAM-AD exhibits significant improvements, and we advocate for more institutions to train SAM for AD contexts. 
4) To further reduce noise interference and retain essential signal features, we design a \textbf{Depth-Guided Wavelet Attention (DGWA)} module that effectively attenuates both high-frequency and low-frequency noises.
5) After fusing point cloud features and image features, we propose {\bf Adaptive Fusion} to further enhance feature robustness and noise resistance through self-attention to re-weight the fused features adaptively. 
We validate RoboFusion's robustness against OOD noise scenarios in KITTI-C and nuScenes-C datasets \cite{Robustness3d}, achieving SOTA performance amid noise, as shown in Fig.~\ref{fig:motivation}.
% We validate our RoboFusion's robustness by simulating OOD noisy scenarios using the KITTI-C and nuScenes-C datasets \cite{Robustness3d}. For example, we simulate OOD noises like rain, snow, fog, strong sunlight in the KITTI validation set while maintaining clean training data, as shown in Fig.~\ref{fig:motivation}. 
% Our %robust framework, 
% RoboFusion achieves SOTA performance on the noisy KITTI-C and nuScenes-C datasets.

% In a nutshell, our contribution is three-fold.
% 1) We propose RoboFusion, a robust multi-modal detection framework utilizing VFMs such as SAM, specifically designed for OOD noise scenarios. 2)

\begin{figure*}[t]
\centering
\includegraphics[width=0.85\textwidth]{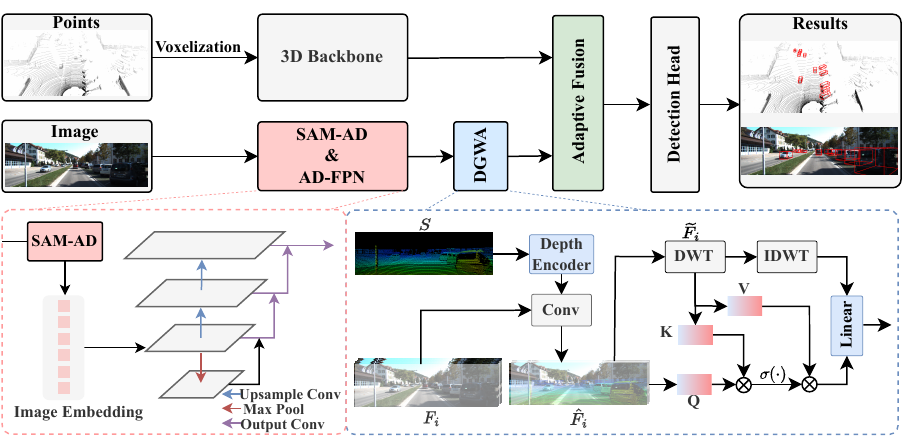}
\caption[ ]{The framework of RoboFusion. The LiDAR branch %almost 
follows the baselines ~\cite{focalconv,transfusion} to generate LiDAR features. In the camera branch, first, we extract robust image features using a highly optimized SAM-AD and acquire multi-scale features using AD-FPN. Second, the sparse depth map $S$ is generated by the raw points and fed into a depth encoder to obtain depth features and fused with multi-scale image features $F_i$ to obtain depth-guided image features $\hat{F}_i$. Then wave attention is used to remove the mutation noise. Finally, adaptive Fusion integrates point cloud features with robust image features with depth information via self-attention mechanism.} 
\label{fig:framework}
\end{figure*}

\begin{figure*}[t]
\centering
\includegraphics[width=0.8\linewidth]{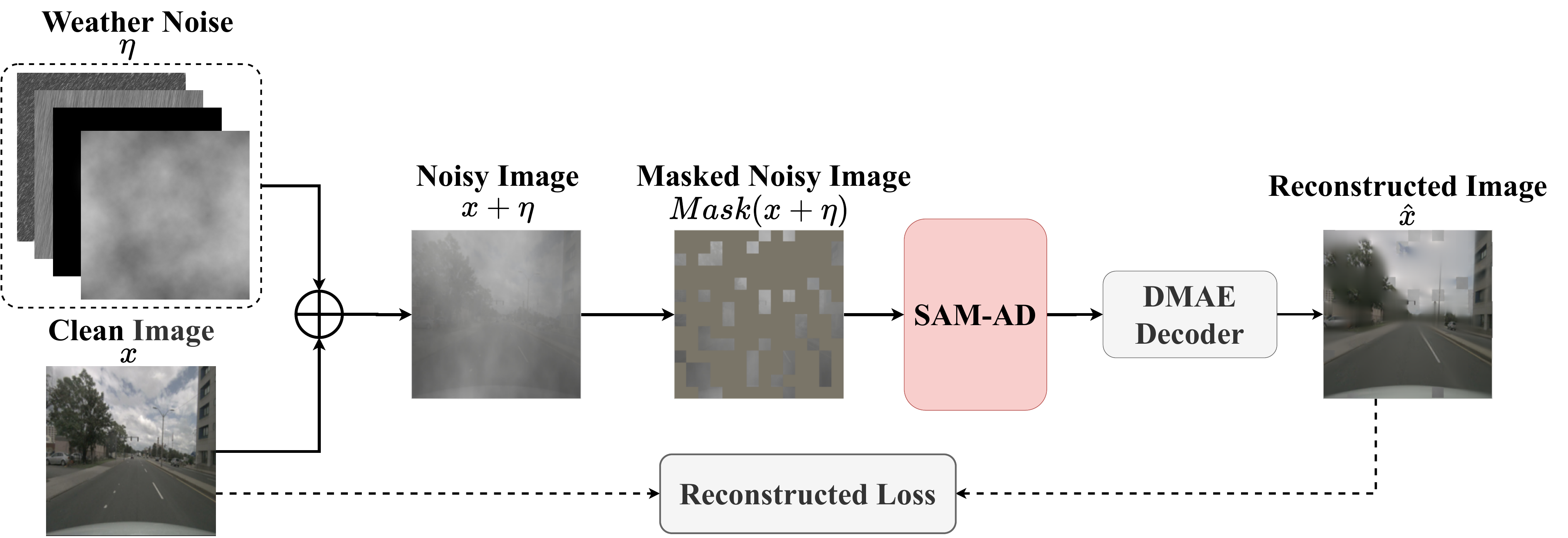}
\caption{An illustration of the pre-training framework. We corrupt a clean image $x$ by $\eta$ which contains multiple weather noises and then randomly masking several patches on a noisy image $x + \eta$ to obtain a masked noisy image $Mask(x+\eta)$. The SAM-AD and DMAE decoder are trained to reconstruct the clean image $\hat{x}$ from $Mask(x+\eta)$. }
\label{fig:pretrain}
\end{figure*}

\begin{figure}[ht] 
\centering 
\includegraphics[width=0.65\linewidth]{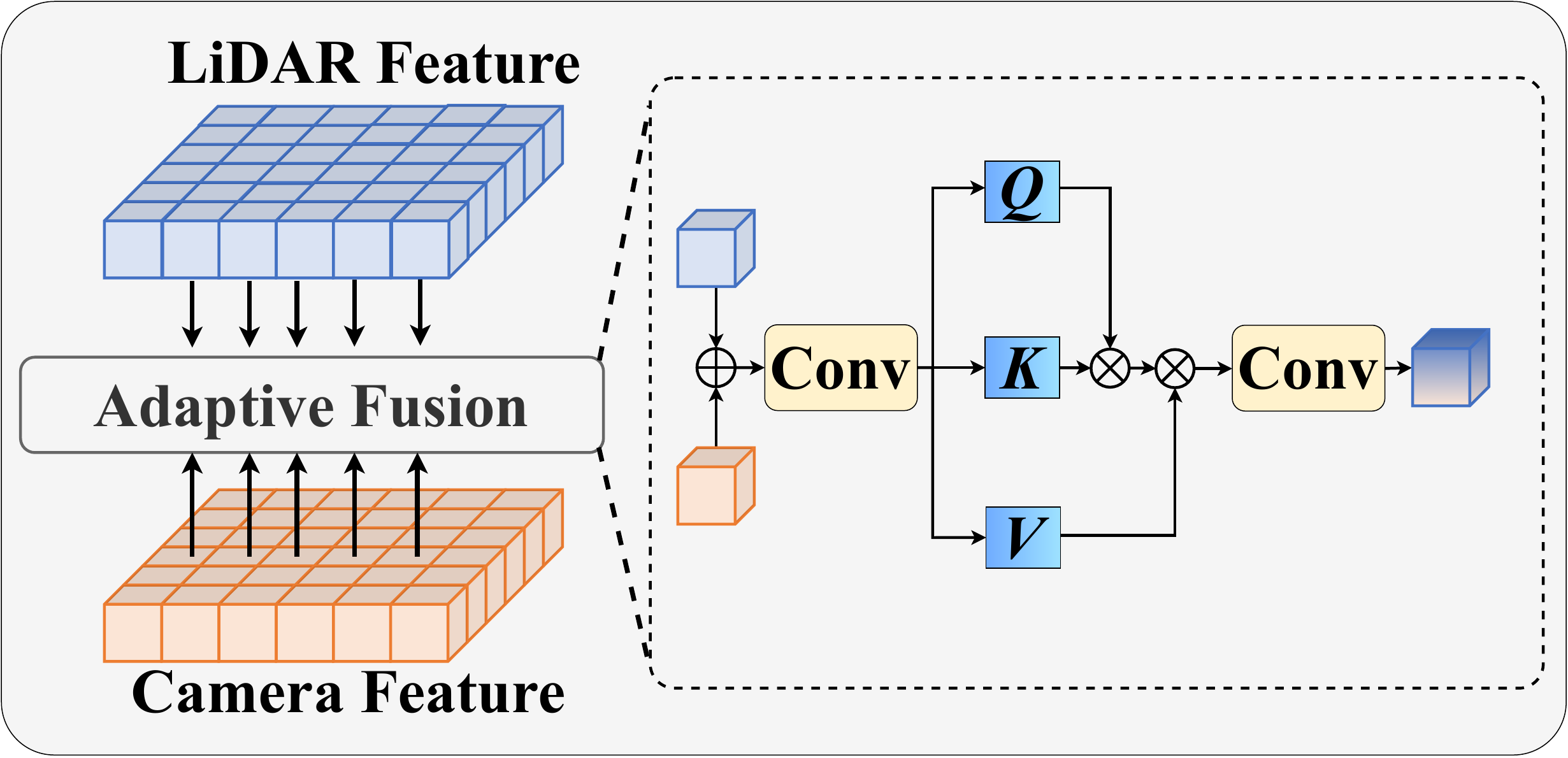} 
\caption{ The architecture of \textbf{Adaptive Fusion}, which involves adaptively re-weighting the fused features using self-attention.} 
% \textbf{(a)} The corruption degree on different modalities when facing different noises. Results from~\cite{3dssd,imvoxelnet,Robustness3d}. 
% \textbf{(b)} An illustrate of Adaptive Fusion. The fused features are re-weighted adaptively by self-attention.
% 传感器破坏不平衡问题。
\label{fig:fusion} 
\end{figure} 

%-------------------------------------------------------------------------
\section{Related Work} 
\subsection{Multi-Modal 3D Object Detection} 
Currently, multi-modal 3D object detection has received considerable attention on popular datasets \cite{kitti,nuscenes}. 
BEVFusion~\cite{bevfusion-mit} fuse multi-modal representations in a unified 3D or BEV space. 
TransFusion~\cite{transfusion} builds a two-stage pipeline where proposals are generated based on LiDAR features and further refined using query image features. 
DeepInteraction~\cite{deepinteraction} and SparseFusion~\cite{sparsefusion} further optimize the camera branch on top of TransFusion. Previous methods are highly optimized to achieve the best performance on clean datasets. However, they ignore common factors in the real world (\textit{e.g.,} bad weather and sensor noise). In this work, we consider a real-world robustness perspective and design a robust multi-modal 3D perception framework, RoboFusion.

\subsection{Visual Foundation Models for 3D Object Detection}
 
Motivated by the success of Large Language Models (LLMs) \cite{gpt4}, VFMs start to be explored in CV community. 
SAM \cite{sam} leverages ViT \cite{vit} to train on the huge SA-1B dataset, containing 11 million samples, which enables SAM to be generalized to many scenes. 
Currently, there have been a few research endeavors aiming at integrating 3D object detectors with SAM. For instance, SAM3D~\cite{sam3d}, as a LiDAR-only method, solely transforms LiDAR's 3D perspective into a BEV (Bird's Eye View) 2D space to harness the generalization capabilities of SAM, yielding sub-optimal performance on `clean' datasets. Another in progress work, 3D-Box-Segment-Anything \footnote{\url{https://github.com/dvlab-research/3D-Box-Segment-Anything}}, tries to utilize SAM for 3D object detection. This indicates the highly attention of SAM like foundation models in 3D scenes in the literature.
Our RoboFusion, as a multi-modal method, gives clear strategies to leverage the generalization capabilities of VFMs to address the OOD noise challenges inherent in existing 3D multi-modal object detection methods.

\section{RoboFusion}
In this section, we present RoboFusion, a framework that harnesses the robustness and generalization capabilities of VFMs such as SAM~\cite{sam} for multi-modal 3D object detection. The overall architecture is depicted in Fig.~\ref{fig:framework} and comprises the following components: 1) \textbf{SAM-AD \& AD-FPN} module which obtains robust multi-scale image features, 
%1) We introduce \textbf{SAM-AD & AD-FPN} to obtain robust multi-scale image features.
%2) We propose the 
2) \textbf{Depth-Guided Wavelet Attention (DGWA)} module which employs wavelet decomposition to denoise depth-guided image features, 
%3)We present the 
3) \textbf{Adaptive Fusion} module which adaptively fuses point cloud features with image features.

% In this section, we present a detailed overview of RoboFusion. First, we showcase the application of VFMs in multi-modal 3D object detection, namely \textbf{SAM-AD}. Secondly, we analyze the drawbacks of using VFMs in the presence of noise and propose the \textbf{depth-guided Wavelet-Attention} (DGWA). Subsequently, we provide the solution after fusing point cloud and image data. Finally, we demonstrate the application of RoboFusion on the strong baselines.

\subsection{SAM-AD \& AD-FPN}
\textbf{Preliminaries.} SAM~\cite{sam}, a VFM, achieves  generalization across diverse scenes due to its extensive training on the large-scale SA-1B dataset—with over 11 million samples and 1 billion high-quality masks. Currently, %VFMs
SAM family \cite{sam,fastsam,mobilesam} primarily support 2D tasks. However, directly extending VFMs like SAM to 3D tasks presents a gap. To address this, we combine SAM with multi-modal 3D models, merging 2D robust feature representations with 3D point cloud features to achieve robust fused features.

\textbf{SAM-AD.} 
To further adapt SAM with AD (autonomous driving) scenarios, we perform pre-training on SAM to obtain SAM-AD. Specifically, we curate an extensive collection of image samples from well-established datasets (\textit{i.e.,} KITTI \cite{kitti} and nuScenes \cite{nuscenes}), forming the foundational AD dataset. 
% Drawing Inspiration from 
% Following DMAE~\cite{dmae}, we devise a pre-training framework for SAM-AD, ensuring that the learned features maintain robustness against additive noise (detail see Appendix). 
Following DMAE \cite{dmae}, we perform pre-training on SAM to obtain SAM-AD in AD scenarios, as shown in Fig.~\ref{fig:pretrain}.
We denote $x$ as a clean image from the AD dataset (\textit{i.e.} KITTI \cite{kitti} and nuScenes \cite{nuscenes}) and $\eta$ as a set of noise images generated by~\cite{Robustness3d} based on $x$. And the noise type and the severity are randomly chosen from the four weather (i.e., rain, snow, fog, and strong sunlight) and the five severities from 1 to 5, respectively. We employ the image encoder of SAM~\cite{sam}
% , FastSAM~\cite{fastsam}, 
, MobileSAM~\cite{mobilesam}
as our encoder while the decoder and the reconstruction loss are the same as DMAE~\cite{dmae}. 
For FastSAM~\cite{fastsam}, we adopt YOLOv8 \footnote{\url{https://github.com/ultralytics/ultralytics }}  to pre-train FastSAM on the AD dataset.
% 我们使用2D检测框架在AD dataset进行预训练。 对于数据增强，我们使用
To avoid overfitting, we use random resizing and cropping as data augmentation. We also set the mask ratio as 0.75 and have trained 400 epochs on 8 NVIDIA A100 GPUs.

\textbf{AD-FPN.} As a promptable segmentation model, SAM has three components: image encoder, prompt encoder and mask decoder. Generally, %it is necessary to generalize image encoder to train VFMs and then to train the decoder. In other words, 
the image encoder can provide high-quality and highly robust image embedding for downstream models, while the mask decoder is only designed to provide decoding services for semantic segmentation. Furthermore, what we require are robust image features rather than the processing of prompting information by the prompt encoder.
% 动机
Therefore, we employ SAM's image encoder to extract robust image features. However, SAM utilizes the ViT series~\cite{vit} as its image encoder, which excludes multi-scale features and provides only high-dimensional low-resolution features. To generate the multi-scale features required for object detection, inspired by~\cite{vitdet}, we design an AD-FPN that offers ViT-based multi-scale features. Specifically, leveraging height-dimensional image embedding with stride 16 (scale=$1/16$) provided by SAM, we produce a series of multi-scale features $F_{ms}$ with stride of $\left \{32, 16, 8, 4 \right \}$. Sequentially, we acquire multi-scale feature $F_i \in \mathbb{R}^{\frac{H}{4} \times \frac{W}{4} \times C_i}$ by integrate $F_{ms}$ in a bottom-up manner similar to FPN~\cite{fpn}. 
% (detail in Appendix)

\subsection{Depth-Guided Wavelet Attention} 
% 动机：与多模态关联起来 
% 可否提出一个问题 为什么深度指引 为什么小波 what why how 
% 用两个图来说明问题 1. 将深度波段分解 2. 将图像分解 
% Although SAM-AD can extract robust image features, our intuition indicates this is insufficient for robust multi-modal perception. 
% In corruption environment, cameras that lost geometric information usually amplify the noise and create negative migration problems. 
Although SAM-AD or SAM has the capability to extract robust image features, the gap between 2D and 3D domains still persists and cameras lacking geometric information in a corrupted environment often amplify noise and give rise to negative migration issues.
To mitigate this problem, we propose the Depth-Guided Wavelet Attention (DGWA) module, which can be split into two steps. %as follows. 
1) A depth-guided network is designed, that adds geometry prior to image features by combining image features and depth features from a point cloud. 2) %We decompose 
The features of an image are decomposed into four wavelet subbands using the Haar wavelet transform~\cite{haar}, then attention mechanism allows to denoise informative features in the subbands. 

Formally, given %a frame of
image features $F_i \in \mathbb{R}^{\frac{H}{4} \times \frac{W}{4} \times C_i}$ and raw points $P \in \mathbb{R} ^ {N, C_p}$ as input. We project $P$ onto the image plane to acquire a sparse depth map $S \in \mathbb{R}^{H \times W \times 2}$. Next, we feed $S$ into the depth encoder $DE(\cdot)$, which consists of several convolution and max pooling blocks, to acquire depth features $F_d \in \mathbb{R} ^ {\frac{H}{4} \times \frac{W}{4} \times C_i}$. Afterward, we leverage convolution encode $(F_i, F_d)$ to acquire depth-guided image features $\hat{F_i} \in \mathbb{R} ^ {\frac{H}{4} \times \frac{W}{4} \times {16}}$, given by 
\begin{equation}
    \hat{F_i}  = Conv(Concat(F_i, DE(S))).
\end{equation}
% where $Conv(\cdot)$ is the 2D convolution operator to encode image feature guided by depth. 

Subsequently, we employ discrete wavelet transform (DWT), a reversible operator, to partition the input $\hat{F_i}$ into four subbands. 
Specifically, we encode the rows and columns of the input separately into one low-frequency band $\widetilde{f}_i^{LL} \in \mathbb{R} ^ {\frac{H}{8} \times \frac{W}{8} \times 4}$  and three high-frequency bands $(\widetilde{f}_i^{LH}, \widetilde{f}_i^{HL}, \widetilde{f}_i^{HH})  \in \mathbb{R} ^ {\frac{H}{8}, \frac{W}{8}, 4}$, with the low-filter $f_L= (\frac{1}{\sqrt{2}}, \frac{1}{\sqrt{2}})$ and the high-filter $f_H = (\frac{1}{\sqrt{2}}, -\frac{1}{\sqrt{2}})$. In this state, the low-frequency band retains coarse-grained information while the high-frequency band retains fine-grained information. In other words, it is easier to capture the mutation signal, so as to filter the noise information. 
We concatenate the four-subband features along channel dimension to acquire wavelet features $\widetilde{F_i} = [\hat{f}_i^{LL}, \hat{f}_i^{LH}, \hat{f}_i^{HL}, \hat{f}_i^{HH}] \in \mathbb{R}^{\frac{H}{8} \times \frac{W}{8} \times 16}$. Next, we perform wave-attention $Att_{\omega}$ to query informative features in the wavelet features. Concretely, we employ $\hat{F}_i$ as a Query and %the wavelet feature
$\widetilde{F_i}$ as a Key/Value given by
\begin{equation}
    F_{att} = Att_{\omega}(\hat{F_i}, \widetilde{F_i}) = \sigma(\frac{\hat{F_i}W^q (\widetilde{F_i}W^k)^T}{\sqrt{C_i}}) \widetilde{F_i}W^v.
\end{equation}
% where $W^q, W^k, W^v$ represent the embedding layer corresponding to $x$ and $\sigma(\cdot)$ represent softmax function. 

Finally, we leverage the IDWT (inverse DWT) to convert $\widetilde{F_i}$ back to $\hat{F_i}$ and 
integrate this converted $\hat{F_i}$ and $F_{att}$ to obtain denoise features $F_{out} \in \mathbb{R} ^ {\frac{H}{16} \times \frac{W}{16} \times 16} $ by 
\begin{equation} 
  F_{out} = MLP(Concat(F_{att}, \hat{F_i})), 
\end{equation} 
where $F_{out}$ preserves informative features and restrains redundant mutation noise in the frequency domain. 
% Sparse Depth的优势 
% Unlike previous methods that require predicting dense depth maps to lift images into pseudo points or voxels, we only leverage sparse depth provided by raw LiDAR, which makes the depth guide more effective. 

\subsection{Adaptive Fusion} 
\label{sec:fusion}
% what why how 
% what 两种数据造成的影响不一样 
% why 用自适应融合 
% how self attention 

% 问题
Following the incorporation of image depth features within the DGWA module, we propose the \textbf{Adaptive Fusion} technique to combine point cloud attributes with robust image features enriched with depth information. Specifically, different types of noise affect LiDAR and images to different degrees, which raises a corruption imbalance problem. Therefore, considering the distinct influences of various noises on LiDAR and camera, we employ self-attention to re-weight the fused features adaptively as shown in Fig.~\ref{fig:fusion}. The corruption degree of modality-specificity is dynamic, and self-attention mechanism allows adaptive re-weighting features to enhance informative features and suppress redundant noise.

\section{Experiments}
% We first introduce  benchmark datasets, including \textbf{KITTI} \cite{kitti}, \textbf{KITTI-C} \cite{Robustness3d}, \textbf{nuScenes} \cite{nuscenes} and \textbf{nuScenes-C} \cite{Robustness3d}, and the experimental setups.  We then compare  RoboFusion with SOTA methods under both clean and noisy %sensor 
% settings. We ablate our method in detail. 
\subsection{Datasets} 
We perform experiments on both the clean public benchmarks (KITTI \cite{kitti} and nuScenes \cite{nuscenes}) and the noisy public benchmarks (KITTI-C\cite{Robustness3d} and nuScenes-C \cite{Robustness3d}).
\subsubsection{KITTI} 
The KITTI dataset provides synchronized LiDAR point clouds and front-view camera images, consists of 3,712 training samples, 3,769 validation samples and 7,518 test samples. The standard evaluation metric for object detection is the mean Average Precision (mAP), computed using recall at 40 positions (R40). 

\subsubsection{nuScenes}
The nuScenes dataset is a large-scale 3D detection benchmark consisting of 700 training scenes, 150 validation scenes, and 150 testing scenes. The data are collected using six multi-view cameras and a 32-channel LiDAR sensor. It includes 360-degree object annotations for 10 object classes. To evaluate the detection performance, the primary metrics used are the mean Average Precision (mAP) and the nuScenes detection score (NDS).

\subsubsection{KITTI-C and nuScenes-C}
In terms of data robustness, \cite{Robustness3d} has designed 27 types of common corruptions for both LiDAR and camera, with the aim of benchmarking the corruption robustness of existing 3D object detectors. \cite{Robustness3d} has established corruption robustness benchmarks \footnote{\url{https://github.com/thu-ml/3D_Corruptions_AD}}, including \textbf{KITTI-C} and \textbf{nuScenes-C}, by synthesizing corruptions on public datasets. Specifically, we utilize \textbf{KITTI-C} and \textbf{nuScenes-C} in our work.  It is worth noting that Ref. \cite{Robustness3d} has only added noise to the validation dataset and kept the train and test datasets clear.

\begin{table}[t]
\scriptsize
\centering
\caption{Comparison with SOTA methods on \textbf{KITTI validation and test} sets for car class with AP of $R_{40}$.}
\addtolength{\tabcolsep}{1.3pt}
\renewcommand\arraystretch{1}
\setlength{\tabcolsep}{1.35mm}
{
\begin{tabular}{c|cccc|cccc}
\toprule
\multirow{3}{*}{Method}  & \multicolumn{4}{c|}{AP$_{3D} (\%)$ (\textit{validation set})}                                                             & \multicolumn{4}{c}{AP$_{3D} (\%)$ (\textit{test set})}                                                            \\ \cmidrule(r){2-9}
&                           \multicolumn{1}{c|}{mAP} 
                        &                           \multicolumn{1}{c|}{Easy}           & \multicolumn{1}{c|}{Mod.}           & Hard           & \multicolumn{1}{c|}{mAP}& \multicolumn{1}{c|}{Easy}           & \multicolumn{1}{c|}{Mod.}           & Hard           \\ 
                        \midrule
Voxel R-CNN                        & \multicolumn{1}{c|}{86.84}   & \multicolumn{1}{c|}{92.38}        & \multicolumn{1}{c|}{85.29}          & 82.86          & \multicolumn{1}{c|}{83.19}    & \multicolumn{1}{c|}{90.90}      & \multicolumn{1}{c|}{81.62}          & 77.06 \\
VFF                        & \multicolumn{1}{c|}{86.91}   & \multicolumn{1}{c|}{92.31}        & \multicolumn{1}{c|}{85.51}          & 82.92          & \multicolumn{1}{c|}{83.62}    & \multicolumn{1}{c|}{89.50}      & \multicolumn{1}{c|}{82.09}          & 79.29 \\
CAT-Det                        & \multicolumn{1}{c|}{83.58}   & \multicolumn{1}{c|}{90.12}        & \multicolumn{1}{c|}{81.46}          & 79.15          & \multicolumn{1}{c|}{82.62}    & \multicolumn{1}{c|}{89.87}      & \multicolumn{1}{c|}{81.32}          & 76.68 \\

% SFD                        & \multicolumn{1}{c|}{89.79}   & \multicolumn{1}{c|}{95.52}        & \multicolumn{1}{c|}{88.27}          & 85.57          & \multicolumn{1}{c|}{84.80}    & \multicolumn{1}{c|}{91.73}      & \multicolumn{1}{c|}{84.76}          & 77.92 \\

% VirConv-L                                 & \multicolumn{1}{c|}{89.30}  & \multicolumn{1}{c|}{93.36}          & \multicolumn{1}{c|}{88.71}          & \multicolumn{1}{c|}{85.83}          & \multicolumn{1}{c|}{85.56}  & \multicolumn{1}{c|}{91.41}          & \multicolumn{1}{c|}{85.05}          & \multicolumn{1}{c}{80.22}                  \\
LoGoNet                                   & \multicolumn{1}{c|}{87.13}  & \multicolumn{1}{c|}{92.04}          & \multicolumn{1}{c|}{85.04}          & \multicolumn{1}{c|}{84.31}         & \multicolumn{1}{c|}{85.87}  & \multicolumn{1}{c|}{91.80}          & \multicolumn{1}{c|}{\textbf{85.06}}          & \multicolumn{1}{c}{\textbf{80.74}}       \\
\midrule
Focals Conv-F  & \multicolumn{1}{c|}{-} & \multicolumn{1}{c|}{-}        & \multicolumn{1}{c|}{-}          & -          & \multicolumn{1}{c|}{83.47} & \multicolumn{1}{c|}{90.55}        & \multicolumn{1}{c|}{82.28}          & 77.59    \\
Baseline*  & \multicolumn{1}{c|}{86.75} & \multicolumn{1}{c|}{92.05}        & \multicolumn{1}{c|}{85.51}          & 82.70          & \multicolumn{1}{c|}{-} & \multicolumn{1}{c|}{-}        & \multicolumn{1}{c|}{-}          & -    \\
\midrule

\rowcolor[HTML]{e0ffff} RoboFusion-L                               & \multicolumn{1}{c|}{\textbf{88.87}}& \multicolumn{1}{c|}{\textbf{93.30}} & \multicolumn{1}{c|}{\textbf{88.04}} & \textbf{85.27}  & \multicolumn{1}{c|}{\textbf{85.58}}& \multicolumn{1}{c|}{91.75} & \multicolumn{1}{c|}{84.08} & 80.71 \\ 
\rowcolor[HTML]{e0ffff} RoboFusion-B                               & \multicolumn{1}{c|}{88.45}& \multicolumn{1}{c|}{93.22} & \multicolumn{1}{c|}{87.87} & 84.27 & \multicolumn{1}{c|}{85.32}& \multicolumn{1}{c|}{\textbf{91.98}} & \multicolumn{1}{c|}{83.76} & 80.23\\ 
\rowcolor[HTML]{e0ffff} RoboFusion-T                               & \multicolumn{1}{c|}{88.08}& \multicolumn{1}{c|}{93.28} & \multicolumn{1}{c|}{87.60} & 83.36 & \multicolumn{1}{c|}{85.09}& \multicolumn{1}{c|}{91.68} & \multicolumn{1}{c|}{83.70} & 79.89  \\ 
\bottomrule
\end{tabular}}
\begin{tablenotes}
\footnotesize
\item[1] $\ast$ denotes our reproduced results based on the officially released codes.
\end{tablenotes}
\label{tab_kitti_test_val}
\end{table}

\begin{table}[]
\scriptsize
\centering
\caption{ Comparison with SOTA methods on \textbf{nuScenes validation and test} sets.}
\renewcommand\arraystretch{1}
\setlength{\tabcolsep}{1.85mm}
{
\begin{tabular}{c|cc|cc|cccc}
    \toprule
\multirow{2}{*}{Method}  &     \multirow{2}{*}{LiDAR }  &\multirow{2}{*}{Camera }   &     \multicolumn{2}{c|}{\textit{validation set}} & \multicolumn{2}{c}{\textit{test set}}   \\
 &&&NDS&mAP &NDS &mAP\\
\midrule 

% VoxelNeXt & VoxelNet & - & 67.1 & 60.0 & 70.0 & 64.5 \\
% CenterPoint & VoxelNet & - & 66.8 & 59.6 & 67.3 & 60.3 \\

FUTR3D & VoxelNet & ResNet-101  & 68.3 & 64.5 & - & -\\
% AutoAlignV2 & VoxelNet & CSPNet & 71.2 & 67.1 & 72.4 & 68.4 \\
BEVFusion-mit& VoxelNet & Swin-T & 71.4 & 68.5 & 72.9 & 70.2 \\
% BEVFusion & VoxelNet & Dual-Swin-T & 72.1 & 69.6 & 73.3 & 71.3 \\
DeepInteraction & VoxelNet & ResNet-50 & 72.6 & 69.9 & 73.4 & 70.8 \\
CMT & VoxelNet & ResNet-50 & 72.9 & 70.3 & 74.1 & 72.0 \\
SparseFusion& VoxelNet & ResNet-50 & 72.8 & 70.4 & 73.8 & 72.0 \\
\midrule
TransFusion & VoxelNet & ResNet-50 & 71.3 & 67.5& 71.6 & 68.9 \\
Baseline* & VoxelNet & ResNet-50 & 70.8 & 67.3 & - & - \\
\midrule
\rowcolor[HTML]{e0ffff}RoboFusion-L & VoxelNet & SAM & 72.1 & 69.9 & 72.0 & 69.9 \\
\rowcolor[HTML]{e0ffff}RoboFusion-B & VoxelNet & FastSAM & 71.9& 69.4& 71.8 & 69.4 \\
\rowcolor[HTML]{e0ffff}RoboFusion-T & VoxelNet & MobileSAM & 71.3 & 69.1 & 71.5 & 69.1 \\
\bottomrule
\end{tabular}
}
\begin{tablenotes}
\footnotesize
\item[1] $\ast$ denotes our reproduced results based on the officially released codes.
\end{tablenotes}
\label{tab_nuScens_test_val}
\end{table}

\begin{table}[t]
\scriptsize
\centering
\caption{ Comparison with SOTA methods on \textbf{KITTI-C validation} set. The results are evaluated based on the car class with AP of $R_{40}$ at moderate difficulty. `S.L.', `D.', `C.O.', and `C.T.' denotes Strong Sunlight, Density, Cutout, and Crosstalk, respectively.}
\renewcommand\arraystretch{1}
\setlength{\tabcolsep}{0.85mm}{
% \begin{tabular}{c|c|cccc|ccccccccc}
\begin{tabular}{c|c|ccccc|ccc}
    \toprule
Rai  &    \multirow{2}{*}{Clean}    & \multicolumn{5}{c|}{\textit{Weather}} & \multicolumn{3}{c}{\textit{Sensor}}  \\
% & & Snow  &    Rain & Fog & Sunlignt & Density & Cutout & Crosstalk  &Gaussian (L)&Uniform (L)&Impulse (L)&Gaussian (C)&Uniform (C)&Impulse (C)\\
& & mAP & Snow  &    Rain & Fog & S.L. & D. & C.O. & C.T.  \\
\midrule

% SECOND$^{\dagger}$ & 81.59 &52.34 & 52.55 & 74.10& 78.32 & 80.18&73.59&80.24&64.90&79.18&81.43&-&-&-\\
SECOND$^{\dagger}$ & 81.59&64.33 &52.34 & 52.55 & 74.10& 78.32 & 80.18&73.59&80.24\\
% PointPillars$^{\dagger}$ & 78.41&36.47& 36.18 & 64.28& 62.28 &76.49&70.28&70.85&74.68&77.31&78.17&-&-&-\\
PointPillars$^{\dagger}$ & 78.41&49.80&36.47& 36.18 & 64.28& 62.28 &76.49&70.28&70.85\\
% PointRCNN$^{\dagger}$ & 80.57 &50.36 & 51.27 & 72.14& 62.78 & 80.35&73.94&71.53&61.20&76.39&79.78&-&-&-\\
PointRCNN$^{\dagger}$ & 80.57&59.14 &50.36 & 51.27 & 72.14& 62.78 & 80.35&73.94&71.53\\
% PV-RCNN$^{\dagger}$ & 84.39 &52.35 & 51.58 & 79.47& 79.91 & 82.79&76.09&82.34&65.11&81.16&82.81&-&-&-\\
PV-RCNN$^{\dagger}$ & 84.39 &65.83&52.35 & 51.58 & 79.47& 79.91 & 82.79&76.09&82.34\\
% SMOKE$^{\dagger}$ & 7.09 &2.47 & 3.94 & 5.63& 6.00 & -&-&-&-&-&-&1.56&2.67&1.83\\
SMOKE$^{\dagger}$ & 7.09 & 4.51 &2.47 & 3.94 & 5.63& 6.00 & -&-&- \\
ImVoxelNet$^{\dagger}$ & 11.49 &3.22&0.22 & 1.24 & 1.34& 10.08 & -&-&-\\
EPNet$^{\dagger}$ & 82.72 &46.21&34.58 & 36.27 & 44.35& 69.65 & 82.09&76.10&82.10\\
Focals Conv-F$^{\dagger}$ & 85.88 &50.40&34.77 & 41.30 & 44.55& 80.97 & 84.95&78.06&\textbf{85.82}\\
LoGoNet*& 85.04 &62.58&51.45 & 55.80 & 67.53& 75.54 & 83.68 & 77.17 & 82.00 \\
% VirConv-L*& 91.95 &57.64 & 58.46 & 78.43& 63.35 & 83.43&76.96&77.50&66.90&81.00&83.85&84.90&84.91&84.91\\
% VirConv-L*& 88.19 &60.00& 50.17 & 50.57& 75.63   & 63.62& 80.70 &75.18&75.67\\
\midrule
% RoboFusion-L& 90.97 &51.17 & 51.57 & 67.53& 75.54 & 83.68&77.17&82.00&61.84&82.94&84.65&84.29&84.45&84.19\\
% RoboFusion-B& 90.97 &51.17 & 51.57 & 67.53& 75.54 & 83.68&77.17&82.00&61.84&82.94&84.65&84.29&84.45&84.19\\
% RoboFusion-T& 90.97 &51.17 & 51.57 & 67.53& 75.54 & 83.68&77.17&82.00&61.84&82.94&84.65&84.29&84.45&84.19\\
\rowcolor[HTML]{e0ffff}RoboFusion-L& \textbf{88.04} &\textbf{85.70}&\textbf{85.29} & \textbf{86.48} & \textbf{85.53}& \textbf{85.50} & \textbf{85.71}&\textbf{83.17}&84.12\\
\rowcolor[HTML]{e0ffff}RoboFusion-B& 87.87&84.70 &84.11 & 85.54 & 84.00& 85.15 & 84.34&81.30&82.45\\
\rowcolor[HTML]{e0ffff}RoboFusion-T& 87.60 &84.60&84.67 & 84.79 & 84.17& 84.75 & 84.11&81.21&83.07\\
\bottomrule
\end{tabular}
}
\begin{tablenotes}
\footnotesize
\item[1] $^{\dagger}$: Results from Ref. \cite{Robustness3d}.
\item[2] * denotes re-implement result.
\end{tablenotes}
\label{tab_KITTI_val_C}
\end{table}

\begin{table}[t]
\scriptsize
\centering
\caption{ Comparison with SOTA methods on \textbf{nuScenes-C validation} set with mAP. `S.L.', `D.', `C.O.', and `C.T.' denotes Strong Sunlight, Density, Cutout, and Crosstalk, respectively.}
\renewcommand\arraystretch{1}
\setlength{\tabcolsep}{0.80mm}
{
% \begin{tabular}{c|c|cccc|ccccccccc}
\begin{tabular}{c|c|ccccc|ccc}
    \toprule
\multirow{2}{*}{Method}  &    \multirow{2}{*}{Clean}    & \multicolumn{5}{c|}{\textit{Weather}} & \multicolumn{3}{c}{\textit{Sensor}}  \\
% & & Snow  &    Rain & Fog & Sunlignt & Density & Cutout & Crosstalk  &Gaussian (L)&Uniform (L)&Impulse (L)&Gaussian (C)&Uniform (C)&Impulse (C)\\
& &mAP & Snow  &    Rain & Fog & S.L. & D. & C.O. & C.T.  \\

\midrule
PointPillars$^{\dagger}$ & 27.69 &25.87&27.57& 27.71 & 24.49& 23.71 &27.27&24.14&25.92\\
 SSN$^{\dagger}$ & 46.65 &43.70 &46.38 & 46.50 &  41.64& 40.28 & 46.14& 40.95&44.08\\
CenterPoint$^{\dagger}$ & 59.28 &52.49 & 55.90 & 56.08 & 43.78& 54.20 &  58.60& 56.28&56.64\\
FCOS3D$^{\dagger}$ & 23.86 &11.44 &2.01 & 13.00 & 13.53& 17.20 & -&-&-\\
PGD$^{\dagger}$ & 23.19 &12.85 &2.30 & 13.51 &  12.83&  22.77 & -&-&-\\
DETR3D$^{\dagger}$ & 34.71 &22.00 &5.08 &  20.39 & 27.89& 34.66 & -&-&-\\
BEVFormer$^{\dagger}$ & 41.65 &26.29 & 5.73 & 24.97 & 32.76&41.68                
 & -&-\\
FUTR3D$^{\dagger}$& 64.17 &55.50 &52.73 & 58.40 & 53.19& 57.70 & 63.72&62.25&62.66\\
TransFusion$^{\dagger}$& 66.38 &58.87 &63.30 & 63.35 &  53.67&  55.14 & 65.77& 63.66&64.67\\
BEVFusion$^{\dagger}$& 68.45 &61.87 &62.84 & 66.13 &  54.10&  64.42 & 67.79 & 66.18&67.32\\
DeepInteraction$^{*}$& 69.90 &62.14 & 62.36 &66.48 & 54.79 &  64.93 & 68.15 & 66.23 &68.12 \\
CMT$^{*}$& \textbf{70.28} &63.46 & 62.56  &61.44  & 66.26 &  63.59 & \textbf{69.65} & 68.70 &68.26 \\
\midrule
\rowcolor[HTML]{e0ffff}RoboFusion-L& 69.91 &\textbf{67.24} &\textbf{67.12} & \textbf{67.58} & \textbf{67.01}& \textbf{67.24} & 69.48&\textbf{69.18}&\textbf{68.68}\\
\rowcolor[HTML]{e0ffff}RoboFusion-B& 69.40 &66.33 &66.07 & 67.01 & 65.54& 66.71 & 69.02&69.01&68.04\\
\rowcolor[HTML]{e0ffff}RoboFusion-T& 69.09 &65.82 &65.96 & 66.45 & 64.34& 66.54 & 68.58&68.20&68.17\\

% PointPillars$^{\dagger}$ & 27.69&27.57& 27.71 & 24.49& 23.71 &27.27&24.14&25.92&19.41&25.60&26.44&-&-&-\\
%  SSN$^{\dagger}$ & 46.65 &46.38 & 46.50 &  41.64& 40.28 & 46.14& 40.95&44.08& 39.16&45.00&45.58& -&-&-\\
% CenterPoint$^{\dagger}$ & 59.28 & 55.90 & 56.08 & 43.78& 54.20 &  58.60& 56.28&56.64&45.79&56.12&57.67& -&-&-\\
% FCOS3D$^{\dagger}$ & 23.86 &2.01 & 13.00 & 13.53& 17.20 & -&-&-&-&-&-&3.96&8.12&3.55\\
% PGD$^{\dagger}$ & 23.19 &2.30 & 13.51 &  12.83&  22.77 & -&-&-&-&-&-&4.33& 8.48& 3.78\\
% DETR3D$^{\dagger}$ & 34.71 &5.08 &  20.39 & 27.89& 34.66 & -&-&-&-&-&-&14.86& 21.49& 14.32\\
% BEVFormer$^{\dagger}$ & 41.65 & 5.73 & 24.97 & 32.76&41.68                
%  & -&-&-&-&-&-&15.04& 23.00&13.99\\
% FUTR3D$^{\dagger}$& 64.17 &52.73 & 58.40 & 53.19& 57.70 & 63.72&62.25&62.66&58.94&63.21&63.43&54.96&57.61&55.16\\
% TransFusion$^{\dagger}$& 66.38 &63.30 & 63.35 &  53.67&  55.14 & 65.77& 63.66&64.67& 55.10& 64.72& 65.51&64.52&65.26&64.37\\
% BEVFusion$^{\dagger}$& 68.45 &62.84 & 66.13 &  54.10&  64.42 & 67.79& 66.18&67.32& 60.64& 66.81& 67.54&64.44&65.81&64.30\\
% \midrule
% RoboFusion-L& 90.97 &51.17 & 51.57 & 67.53& 75.54 & 83.68&77.17&82.00&61.84&82.94&84.65&84.29&84.45&84.19\\
% RoboFusion-B& 90.97 &51.17 & 51.57 & 67.53& 75.54 & 83.68&77.17&82.00&61.84&82.94&84.65&84.29&84.45&84.19\\
% RoboFusion-T& 90.97 &51.17 & 51.57 & 67.53& 75.54 & 83.68&77.17&82.00&61.84&82.94&84.65&84.29&84.45&84.19\\

\bottomrule
  \end{tabular}
}
  \begin{tablenotes}
\footnotesize
\item[1] $^{\dagger}$: Results from Ref. \cite{Robustness3d}.
\item[2] * denotes re-implement result.

\end{tablenotes}

  \label{tab_nuscenes_val_C}

\end{table}

\begin{table}[]
\scriptsize
\centering
\caption[ ]{Performance of different VFMs on RoboFusion. `RCE' denotes Relative Corruption Error \cite{Robustness3d}. `mAP (Weather)' denotes the average value across four types of weather corruptions, Snow, Rain, Fog, and Strong Sunlight.}
\addtolength{\tabcolsep}{1.3pt}

\renewcommand\arraystretch{1}
\setlength{\tabcolsep}{0.5mm}{

\begin{tabular}{c|c|c|c|c|c}
\toprule
Method & Model Size & FPS (A100) & mAP (Weather ) & mAP (Clean) & RCE (\%) \\
\midrule
RoboFusion-L & 97.54M & 3.1 & 67.24 & 69.91 & 0.04\\
RoboFusion-B & 81.01M & 3.5 & 66.33 & 69.40 & 0.04\\
RoboFusion-T & 13.94M & 6.0 & 65.82 & 69.09 & 0.05\\
\midrule
DeepInteraction&57.82M&4.9& 62.14&69.90&0.10\\
TransFusion&36.96M &6.2&58.37&66.38&0.12\\
\bottomrule
\end{tabular}}
\label{tab_kitti_modelsize_fps}
\end{table}

\subsection{Experimental Settings}
\subsubsection{Network Architecture.}
Our RoboFusion consists of three variants: RoboFusion-L, RoboFusion-B, and RoboFusion-T, which utilize the models SAM-B \cite{sam}, FastSAM \cite{fastsam}, and MobileSAM \cite{mobilesam}, respectively. 
It is noteworthy that due to the convolutional operations of FastSAM in RoboFusion-B which is capable of generating multi-scale features, the AD-FPN module is not employed.
Since KITTI and nuScenes are distinct datasets with varying evaluation metrics and characteristics, we provide a detailed description of our RoboFusion settings for each dataset. % in the following section. 

\textbf{RoboFusion in KITTI and KITTI-C.} We validate our RoboFusion on the KITTI  dataset using  Focals Conv  \cite{focalconv} as the baseline. The input voxel size is set to (0.05m, 0.05m, 0.1m), with anchor sizes for cars at [3.9, 1.6, 1.56] and anchor rotations at [0, 1.57]. We adopt the same data augmentation solution as Focals Conv-F.

\textbf{RoboFusion with nuScenes and nuScenes-C.} We validate our RoboFusion on the nuScenes  dataset using  TransFusion \cite{transfusion} as the baseline. The detection range for the X and Y axis is set at [-54m, 54m] and [-5m, 3m] for the Z axis. The input voxel size is set at (0.075m, 0.075m, 0.2m), and the maximum number of point clouds contained in each voxel is set to 10. 
It is noteworthy that the Adaptive Fusion module is applied exclusively to Focals Conv rather than TransFusion, while TransFusion uses its own fusion module.% utilizing the fusion module from TransFusion.

\subsubsection{Training and Testing Details.}
Our RoboFusion is meticulously trained from scratch using the Adam optimizer and incorporates several foundation models as image encoders including SAM, FastSAM and MobileSAM. To enable effective training on the KITTI and nuScenes datasets, we utilize 8 NVIDIA A100 GPUs for network training. Additionally, the runtime is evaluated on an NVIDIA A100 GPU. Specifically, for KITTI, our RoboFusion based on Focals Conv\cite{focalconv} 
involves training for 80 epochs. For nuScenes, our RoboFusion based on %CenterPoint  \cite{centerpoint} and 
TransFusion 
\cite{transfusion} 
has 20 epochs of training.
During the model inference stage, we employ a non-maximal suppression (NMS) operation in the Region Proposal Network (RPN) with an IoU threshold of 0.7. We select the top 100 region proposals to serve as inputs for the detection head. After refinement, we apply NMS again with an IoU threshold of 0.1 to eliminate redundant predictions.
For additional details regarding our method, please refer to OpenPCDet \footnote{\url{https://github.com/open-mmlab/OpenPCDet}} .

\subsection{Comparing with state-of-the-art}
We conduct evaluations on the clean datasets KITTI and nuScenes, as well as the noisy datasets KITTI-C and nuScenes-C. While SOTA methods are primarily focused on achieving high accuracy, we place greater emphasis on the robustness and generalization of the methods. These factors are crucial for the practical deployment of 3D object detection in AD scenarios, making the evaluation on the noisy datasets more important in our perspective.

\subsubsection{Results on the clean benchmark.}

As shown in Table \ref{tab_kitti_test_val}, we compare our RoboFusion with SOTA methods, including Voxel R-CNN \cite{voxelrcnn}, VFF \cite{vff}, CAT-Det\cite{cat-det} , Focals Conv-F \cite{focalconv}, and LoGoNet \cite{logonet} on the KITTI validation and test sets.  As shown in Table \ref{tab_nuScens_test_val}, we also compare our RoboFusion with SOTA methods, including  FUTR3D \cite{chen2023futr3d}, TransFusion \cite{transfusion}, BEVFusion \cite{bevfusion-mit}, DeeepInteraction \cite{deepinteraction}, CMT \cite{cmt} and SparseFusion \cite{sparsefusion}, on the nuScenes test and validation sets. 
Our RoboFusion has achieved SOTA performance on the clean benchmarks (KITTI and nuScenes).
% Due to the adoption of different paradigms, we have little gaps compare to the 1st rank VirConv-L on the KITTI leaderboard and leading CMT on the nuScenes leaderboard. However, our RoboFusion outperforms other SOTA methods like CMT and LoGoNet in terms of robustness and generalization, especially in noisy scenarios, as demonstrated in Tables \ref{tab_KITTI_val_C} and \ref{tab_nuscenes_val_C}.

\subsubsection{Results on the noisy benchmark.}
In the real-world AD scenarios, the distribution of data often differs from that of training or testing data, as shown in Fig.~\ref{fig:motivation} (a). 
Specifically, Ref. \cite{Robustness3d} provides a novel noisy benchmark that includes KITTI-C and nuScenes-C, which we primarily use to evaluate the weather and sensor noise corruptions, including rain, snow, fog, and strong sunlight, density, cutout, and so on. In addition, comparisons of our RoboFusion with SOTA methods in other settings are presented in the Appendix \footnote{\url{https://arxiv.org/abs/2401.03907}}.

As shown in Table \ref{tab_KITTI_val_C}, SOTA methods, including SECOND \cite{SECOND}, PointPillars \cite{pointpillars}, PointRCNN \cite{pointrcnn}, PV-RCNN \cite{pvrcnn}, SMOKE \cite{liu2020smoke}, ImVoxelNet \cite{imvoxelnet}, EpNet\cite{epnet}, Focals Conv-F\cite{focalconv}, and LoGoNet\cite{logonet}, experience a significant decrease in performance on the noisy scenarios, particularly for weather conditions such as snow and rain. It can be attributed to the fact that the `clean' KITTI  dataset does not include examples %of learning 
in snowy or rainy weather. % which is the underlying reason for this drop in performance. 
On the other hand, VFMs like %our 
SAM-AD have been trained on a diverse range of data and exhibit robustness and generalization to OOD scenarios, leading to higher performance on our RoboFusion metric. Furthermore, multi-modal methods like  LoGoNet, and Focals Conv-F demonstrate better robustness and generalization in sensor noise scenarios, while LiDAR-only methods like PV-RCNN \cite{pvrcnn} are more robust in weather noise scenarios. This observation motivates our research on adaptive fusion schemes for point cloud and image features. Overall, in the KITTI-C \cite{Robustness3d} dataset, our RoboFusion's performance is nearly on par with the clean scene, indicating high level of robustness and generalization. 

As shown in Table \ref{tab_nuscenes_val_C}, SOTA methods including PointPillars \cite{pointpillars}, SSN \cite{ssn}, CenterPoint \cite{centerpoint}, FCOS3D \cite{fcos3d}, PGD \cite{pgd}, DETR3D \cite{detr3d}, BEVFormer \cite{li2022bevformer}, FUTR3D \cite{chen2023futr3d}, TransFusion \cite{transfusion}, BEVFusion\cite{bevfusion-mit}, DeepInteraction \cite{deepinteraction} and CMT \cite{cmt} in nuScenes-C show relatively higher robustness than in KITTI-C when faced with weather noise. However, %in foggy scenarios, 
BEVFusion performs well in the presence of snow, rain, and strong sunlight noise but experiences a significant performance drop in foggy scenarios. In contrast, our method exhibits strong robustness and generalization in both weather and sensor noise scenarios in nuScenes-C.

\begin{table}[t]
\scriptsize
\centering
\caption{Impacts of different SAM usages on \textbf{KITTI} and \textbf{KITTI-C validation} sets for car class with AP of R$_{40}$. `S.L.' denotes Strong Sunlight.
}
\addtolength{\tabcolsep}{1.3pt}

\renewcommand\arraystretch{1}
\setlength{\tabcolsep}{1.58mm}{

\begin{tabular}{c|cccc|cccc}
\toprule
\multirow{2}{*}{Solution}  & \multicolumn{4}{c|}{AP$_{3D} (\%)$}                                                             & \multicolumn{4}{c}{AP$_{Weather}(\%)$}                                                            \\ \cmidrule(r){2-9}
&                           \multicolumn{1}{c|}{mAP} 
                        &                           \multicolumn{1}{c|}{Easy}           & \multicolumn{1}{c|}{Mod.}           & Hard           & \multicolumn{1}{c|}{Snow}& \multicolumn{1}{c|}{Rain}           & \multicolumn{1}{c|}{Fog}           & S.L.           \\ 
                        \midrule
% RoboFusion-B(offline) 
Offline
& \multicolumn{1}{c|}{80.41}   & \multicolumn{1}{c|}{88.76}        & \multicolumn{1}{c|}{77.38}          & 75.11          & \multicolumn{1}{c|}{-}    & \multicolumn{1}{c|}{-}      & \multicolumn{1}{c|}{-}          & - \\
% RoboFusion-B(no optim) 
No optim
& \multicolumn{1}{c|}{86.45}   & \multicolumn{1}{c|}{91.86}        & \multicolumn{1}{c|}{84.80}          & 82.71          & \multicolumn{1}{c|}{45.11}    & \multicolumn{1}{c|}{47.77}      & \multicolumn{1}{c|}{63.10}          & 79.21 \\
% RoboFusion-B(optim) 
Optim
& \multicolumn{1}{c|}{88.00}   & \multicolumn{1}{c|}{92.41}        & \multicolumn{1}{c|}{86.77}          & 84.81          & \multicolumn{1}{c|}{57.43}    & \multicolumn{1}{c|}{54.27}      & \multicolumn{1}{c|}{68.81}          & 82.07 \\
\bottomrule
\end{tabular}}

\label{tab_abliation_1}
\end{table}
%

%消融表格二
\begin{table}[t]
\scriptsize
\centering
\caption{Influence of pre-training on SAM at \textbf{KITTI-C validation} set for car class with AP of R$_{40}$ at moderate difficulty. `S.L.', `D.', `C.O.', and `C.T.' denotes Strong Sunlight, Density, Cutout, and Crosstalk, respectively.}
\renewcommand\arraystretch{1}
\setlength{\tabcolsep}{2.1mm}{
% \begin{tabular}{c|c|cccc|ccccccccc}
\begin{tabular}{c|cccc|ccc}
    \toprule
\multirow{2}{*}{VFM}   & \multicolumn{4}{c|}{\textit{Weather}} & \multicolumn{3}{c}{\textit{Sensor}}  \\
% & & Snow  &    Rain & Fog & Sunlignt & Density & Cutout & Crosstalk  &Gaussian (L)&Uniform (L)&Impulse (L)&Gaussian (C)&Uniform (C)&Impulse (C)\\
& Snow  &    Rain & Fog & S.L. & D. & C.O. & C.T.  \\
\midrule
  SAM& 57.43    & 54.27      & 68.81          & 82.07  & 84.21 & 83.04 & 84.06\\
SAM-AD & 80.68 & 81.68 & 81.67 & 83.48 & 84.71& 84.17& 84.12\\
\bottomrule
  \end{tabular}
}

\label{tab_abliation_2}
\end{table}

% 消融表格三
\begin{table}[t]
\scriptsize
\centering
\caption{Roles of SAM3DFusion-L modules on \textbf{KITTI-C validation} set for car class with AP of R$_{40}$ at moderate difficulty. `A.F.' denotes \textbf{Adaptive Fusion} module. `S.L.' denotes strong sunlight.}
\renewcommand\arraystretch{0.8}
\setlength{\tabcolsep}{0.55mm}{
% \begin{tabular}{c|c|cccc|ccccccccc}
\vspace{-1.1em}
\begin{tabular}{c|cccc|ccccc}
    \toprule
Method & SAM-AD & AD-FPN  &  DGWA & A.F. & Snow &  Rain & Fog & S.L. & FPS(A100) \\
\midrule
a) & &  & & & 34.77 & 41.30 & 44.55 & 80.97& 10.8\\
b) &\checkmark &  & & & 80.68 & 81.68 & 81.67 & 83.48&4.0\\
c) &\checkmark &\checkmark  & & & 82.32 & 83.60 & 82.39& 83.98&3.6\\
d) &\checkmark &\checkmark  &\checkmark & & 83.99 & 85.63 & 84.01 & 84.81&3.4\\
\rowcolor[HTML]{e0ffff}e) &\checkmark &\checkmark  &\checkmark &\checkmark & 85.29 & 86.48 & 85.53 & 85.50& 3.1\\
\bottomrule
  \end{tabular}
}
\label{tab_abliation_3}
\end{table}
% \begin{table}[htp]
% \scriptsize
% \centering
% \caption{Roles of RoboFusion modules on \textbf{KITTI-C validation} set for car class with AP of R$_{40}$ at moderate difficulty. `A.F.' denotes \textbf{Adaptive Fusion} module. `S.L.' denotes strong sunlight.}
% \renewcommand\arraystretch{0.8}
% \setlength{\tabcolsep}{1.25mm}{
% % \begin{tabular}{c|c|cccc|ccccccccc}
% \begin{tabular}{c|cccc|cccc}
%     \toprule
% Method & SAM-AD & AD-FPN  &  DGWA & A.F. & Snow &  Rain & Fog & S.L.  \\
% \midrule
% a) & &  & & & 34.77 & 41.30 & 44.55 & 80.97 \\
% b) &\checkmark &  & & & 80.68 & 81.68 & 81.67 & 83.48\\
% c) &\checkmark &\checkmark  & & & 82.32 & 83.60 & 82.39& 83.98\\
% d) &\checkmark &\checkmark  &\checkmark & & 83.99 & 85.63 & 84.01 & 84.81\\
% \rowcolor[HTML]{e0ffff}e) &\checkmark &\checkmark  &\checkmark &\checkmark & 85.29 & 86.48 & 85.53 & 85.50\\
% \bottomrule
%   \end{tabular}
% }
% \label{tab_abliation_3}
% \end{table}

\subsection{Ablation Study}
\subsubsection{Performance of Different VFMs on RoboFusion.}
In order to analyze the noise robustness and FPS performance of different-sized VFMs, SAM, FastSAM and MobileSAM,  we conduct comparative experiments  of RoboFusion-L, RoboFusion-B and RoboFusion-T with SOTA methods, DeepInteraction \cite{deepinteraction} and TransFusion \cite{transfusion}, on the nuScenes-C \cite{Robustness3d} validation set, as shown in Table \ref{tab_kitti_modelsize_fps}. Specifically, our RoboFusion exhibits remarkable robustness to weather noise scenarios. Furthermore, our RoboFusion-T has a similar FPS to TransFusion \cite{transfusion}. Overall, we have presented a viable application of SAM in 3D object detection tasks.
% providing a reasonable scheme for its adoption. 
\subsubsection{Impacts of Different SAM usages.}
As shown in Table \ref{tab_abliation_1}, our RoboFusion-L is experimented upon. Specifically, the first row is the offline usage, which involves loading pre-saved image features during training. It implies that certain online data augmentation cannot be utilized. The second (No optim) and the third (Optim) rows are online usages, where the former omits fine-tuning and keeps the model parameters fixed, the latter follows fine-tuning and updating. 
Therefore, offline usage perform worse than online usages. Additionally, fine-tuning the weights of SAM has demonstrated superior performance, resulting in a performance improvement in the presence of snow, rain, and fog noise scenarios.

\subsubsection{Influence of Pre-training on SAM.}
As shown in Table \ref{tab_abliation_2}, to investigate the scientific value of pre-trained VFMs like SAM, FastSAM, and MobileSAM in AD scenarios, we conduct our RoboFusion-L with SAM evaluation on SAM and SAM-AD. Through pre-training, SAM-AD has gained a better understanding of AD scenarios than the original SAM. The pre-training strategy effectively improves the performance of our RoboFusion, demonstrating a significant improvement in the snow, rain, and fog noise scenarios. 

\subsubsection{Roles of Different Modules in RoboFusion.}
As shown in Table \ref{tab_abliation_3}, we present ablation experiments for different modules of our RoboFusion-L, built upon SAM-AD, including AD-FPN, DGWA, and Adaptive Fusion. Leveraging the strong capabilities of SAM-AD in AD scenarios, SAM-AD has a significant improvement from baseline Focals Conv \cite{focalconv} (34.77\%, 41.30\%, 44.55\%, 80.97\%) to (80.68\%, 81.68\%, 81.67\%, 83.48\%). Subsequently, AD-FPN, DGWA, and Adaptive Fusion achieve even higher performance on the foundation of SAM-AD. This further highlights the substantial contributions of diverse modules within our RoboFusion framework in addressing OOD noise scenarios in AD.
% We conduct ablation studies to explore the performance of different modules, as shown in Table \ref{tab_abliation_3}.

\section{Conclusions}
In this work, we propose a robust framework RoboFusion to enhance the robustness and generalization of multi-modal 3D object detectors using VFMs like SAM, FastSAM, and MobileSAM. Specifically, we pre-train SAM for AD scenarios, yielding SAM-AD. To align SAM or SAM-AD with multi-modal 3D object detectors, we introduce AD-FPN for feature upsampling. To further mitigate noise and weather interference, we apply wavelet decomposition for depth-guided image denoising. Subsequently, we utilize self-attention mechanisms to adaptively reweight fused features, enhancing informative attributes and suppressing excess noises. Extensive experiments demonstrate that our RoboFusion effectively integrates VFMs to boost feature robustness and address OOD noise challenges. We anticipate this work to lay a strong foundation for future research on building robust and dependable foundation AD models.

\textbf{Limitation and Future Work.} %There are some limitations on our RoboFusion, 
First, RoboFusion has a heavy reliance on the representation capability of VFMs. This raises the baseline models' generalization ability, but increases their complexities. % of the baseline models. %while a commom way is to leverage complex model designs to enhance robustness. 
Second, the inference speed of RoboFusion-L and RoboFusion-B is relatively slow due to the limitations of SAM and FastSAM. However, the inference speed of RoboFusion-T is competitive with some SOTA methods (e.g. TransFusion) %\cite{transfusion}) 
without VFMs. % on inference speed. 
In the future, for improving the real-time application ability of VFMs, we will attempt to incorporate SAM only in the training phase to guide a fast-speed student model, meanwhile explore more noise scenarios.
\section*{Acknowledgments}
This work was supported by 
the Fundamental Research Funds for the Central Universities (2023YJS019), 
the National Key R\&D Program of China (2018AAA0100302).
\appendix
\section {Appendix}

\subsection{Broader Impacts}
\label{sec:boardimpacts}
% \section{A.1. Broader Impacts}
Our work aims to develop a robust framework to address out-of-distribution (OOD) noise scenarios in autonomous driving (AD). To the best of our knowledge, RoboFusion is the first method that leverages the generalization capabilities of visual foundation models (VFMs) like SAM \cite{sam}, FastSAM \cite{fastsam}, and MobileSAM \cite{mobilesam} for multi-modal 3D object detection. Although existing multi-model 3D object detection methods achieve the state-of-the-art (SOTA) performance of `clean' datasets, they overlook the robustness of real-world scenarios\cite{song2024robustness}. Therefore, we believe it is valuable to combine VFMs and multi-modal 3D object detection to mitigate the impact of OOD noise scenarios.

\subsection{More Results}
\label{sec:additionalresult}
% \subsubsection{More Results on the KITTI validation set.}
% We provide the results of the KITTI validation dataset to better present the detection performance of our RoboFusion, as shown in Table \ref{tab_kitti_val}.
% Compared to SOTA methods, our method achieves better performance. This is particularly evidenced by our version of RoboFusion-B, where it attains a performance of 85.27\%  on the KITTI dataset's hard level. Numerous challenging objects in the hard subset of KITTI underscore our RoboFusion's robustness and generalization in tackling difficult object recognition tasks.

\subsubsection{Specific classes AP on the nuScenes-C validation set.}

As shown in Table \ref{tab_nus_val_weather_noise}, we present a comparison of Specific classes AP between TransFusion and our RoboFusion-L on the nuScenes-C validation set, encompassing scenarios with snow, rain, fog, and strong sunlight noise. It is evident from the results that RoboFusion-L exhibits superior performance.

\begin{table}[htp]
% \small% 设置字体大小命令由小到大依次为 \tiny \scriptsize \footnotesize \small
% \footnotesize
% \scriptsize
\centering
\addtolength{\tabcolsep}{0.1pt}
\caption{Comparison with TransFusion on nuScenes validation \textcolor{blue}{`Snow, Rain, Fog, and Strong Sunlight'} noisy scenarios. `T.F.', `R.F.', `S.L.', `C.V.', `Motor.', `Ped.', and `T.C.' are short for \textcolor{red}{TransFusion}, \textcolor{red}{RoboFusion-L}, Strong Sunlight, construction vehicle, motorcycle, pedestrian, and traffic cone, respectively.  }
\vspace{-0.55em} 
  \renewcommand\arraystretch{1}
  \tabcolsep=0.5mm %%%%%%%%%
  \resizebox{\linewidth}{!}{
  %\begin{tabular*}{\linewidth} {@{}@{\extracolsep{\fill}}!{\color{white}\vline}l|c|c|c|c|c|c|c|c|c|c|c|c @{}}
  \begin{tabular}{ll|c|cc ccccc ccccc }
    \toprule
        && mAP  & Car  & Truck & C.V. & Bus  & Trailer & Barrier & Motor. & Bike & Ped. & T.C. \\ 
\midrule

\multirow{3}{*}{\rotatebox[origin=c]{90}{\textcolor{blue}{Snow}} } 
& T.F.& 63.30 &84.55 & 58.41& 25.50& 62.31& 56.00& 70.19& 69.98& 43.69& 84.24& 78.16\\ 
&R.F. & 67.12 & 87.21& 60.88& 29.47& 67.45& 58.99& 75.12& 71.45& 48.28& 86.23& 86.12\\
&
&\textit{\fontsize{8}{0}\selectfont\textcolor{red}{+3.82}}
&
&
&
&\textit{\fontsize{8}{0}\selectfont\textcolor{red}{+5.14}}
&
&\textit{\fontsize{8}{0}\selectfont\textcolor{red}{+4.93}}
&
&\textit{\fontsize{8}{0}\selectfont\textcolor{red}{+4.59}}
&
&\textit{\fontsize{8}{0}\selectfont\textcolor{red}{+7.96}}
\\
\midrule

\multirow{3}{*}{\rotatebox[origin=c]{90}{\textcolor{blue}{Rain}} } 
& T.F.& 63.35  &85.37 & 56.87& 25.12& 64.65& 55.10& 71.99& 68.21& 44.13& 83.87& 78.14\\ 
&R.F.&  67.58 & 86.79& 60.44& 30.21& 65.41& 58.12& 75.47& 71.39& 50.87& 88.91& 88.23
\\
&
&\textit{\fontsize{8}{0}\selectfont\textcolor{red}{+4.23}}
&
&
&\textit{\fontsize{8}{0}\selectfont\textcolor{red}{+5.09}}
&
&
&
&
&\textit{\fontsize{8}{0}\selectfont\textcolor{red}{+6.74}}
&\textit{\fontsize{8}{0}\selectfont\textcolor{red}{+5.04}}
&\textit{\fontsize{8}{0}\selectfont\textcolor{red}{+10.09}}
\\
\midrule

\multirow{3}{*}{\rotatebox[origin=c]{90}{\textcolor{blue}{Fog}} } & T.F.& 53.67 &80.23& 48.51& 18.04& 50.69& 53.03& 62.24& 54.53& 25.27& 80.63& 66.48\\ 
&R.F.&67.01& 87.56& 59.03& 29.36& 66.10& 57.23& 74.33& 72.01& 49.50& 87.08& 87.91
\\
&
&\textit{\fontsize{8}{0}\selectfont\textcolor{red}{+13.34}}
&
&\textit{\fontsize{8}{0}\selectfont\textcolor{red}{+10.52}}
&\textit{\fontsize{8}{0}\selectfont\textcolor{red}{+11.32}}
&\textit{\fontsize{8}{0}\selectfont\textcolor{red}{+15.41}}
&
&\textit{\fontsize{8}{0}\selectfont\textcolor{red}{+12.09}}
&\textit{\fontsize{8}{0}\selectfont\textcolor{red}{+17.48}}
&\textit{\fontsize{8}{0}\selectfont\textcolor{red}{+24.23}}
&
&\textit{\fontsize{8}{0}\selectfont\textcolor{red}{+21.43}}
\\
\midrule

\multirow{3}{*}{\rotatebox[origin=c]{90}{\textcolor{blue}{S.L.}} } 
& T.F.& 55.14 &81.99& 48.07& 19.78& 51.09& 52.57& 63.68& 55.09& 26.98& 82.68& 69.49\\ 
& R.F.& 67.24 &87.67& 57.74& 31.00& 64.29& 58.94& 75.23& 70.23& 50.82& 88.70& 87.82\\
&
&\textit{\fontsize{8}{0}\selectfont\textcolor{red}{+12.10}}
&
&
&\textit{\fontsize{8}{0}\selectfont\textcolor{red}{+11.22}}
&\textit{\fontsize{8}{0}\selectfont\textcolor{red}{+13.20}}
&
&\textit{\fontsize{8}{0}\selectfont\textcolor{red}{+11.55}}
&\textit{\fontsize{8}{0}\selectfont\textcolor{red}{+15.14}}
&\textit{\fontsize{8}{0}\selectfont\textcolor{red}{+23.30}}
&
&\textit{\fontsize{8}{0}\selectfont\textcolor{red}{+18.33}}
\\
\bottomrule
\end{tabular} }
\label{tab_nus_val_weather_noise}
\end{table}

\subsubsection{Roles of Different Modules in RoboFusion.}
To assess the roles of different modules in RoboFusion, we conduct an ablation study on the original SAM rather than SAM-AD, as shown in Table \ref{tab_abliation}, where
a) is the results of the baseline \cite{focalconv}, b)-e) shows the performance of our RoboFusion-L under different modules. According to Table \ref{tab_abliation}, SAM and AD-FPN modules significantly improve the performance in OOD noisy  scenarios. It is worth noticing that DGWA module significantly improves the performance, especially in snow noisy scenarios. 
By Table \ref{tab_abliation}, the impact of fog noise on point clouds is relatively minor. But, using A.F. (Adaptive Fusion) module to dynamically aggregate point cloud features and image features exhibits significant enhancements in fog-noise scenarios.

\begin{table}[h!t]
\scriptsize
\centering
\caption{Roles of RoboFusion modules on \textbf{KITTI-C validation} set for car class with AP of R$_{40}$ at moderate difficulty. `A.F.' denotes \textbf{Adaptive Fusion} module. `S.L.' denotes Strong Sunlight.}
\label{tab_abliation}
\renewcommand\arraystretch{1.3}
\setlength{\tabcolsep}{1.55mm}{
% \begin{tabular}{c|c|cccc|ccccccccc}
\begin{tabular}{c|cccc|cccc}
\toprule
Method & SAM & AD-FPN  &  DGWA & A.F. & Snow &  Rain & Fog & S.L.  \\
\midrule
a) & &  & & & 34.77 & 41.30 & 44.55 & 80.97 \\
b) &\checkmark &  & & & 57.43    & 54.27      & 68.81          & 82.07\\
%c) &\checkmark &\checkmark  & & & 50.69 & 54.03 & 55.86& 82.51\\
c) &\checkmark &\checkmark  & & & 59.81 & 56.59 & 69.68& 83.20\\
d) &\checkmark &\checkmark  &\checkmark & & 66.45 & 58.11 & 70.53 & 84.01\\
e) &\checkmark &\checkmark  &\checkmark &\checkmark & 68.47 & 59.07 & 74.38 & 84.07\\
\bottomrule
\end{tabular}
}
\end{table}
% -----------kitti-c---------------------
\begin{table*}[h!t]
\scriptsize
\caption[ ]{Comparison with SOTA methods on \textbf{KITTI-C validation} set. The results are evaluated based on the 
\textbf{car} class with AP of $R_{40}$ at \textbf{moderate} difficulty. The best one is highlighted in \textbf{bold}.`S.L.' denote Strong Sunlight. `RCE' denotes Relative Corruption Error from Ref.\cite{Robustness3d}. 
% `P.P.' refers to PointPillars~\cite{pointpillars}, `P.R.' refers to PointRCNN~\cite{pointrcnn}, `PV.R.' refers to PV-RCNN~\cite{pvrcnn}, `I.V.N.' refers to ImVoxelNet~\cite{imvoxelnet}, `F.C.' refers to Focals Conv~\cite{focalconv}, `L.G.N.' refers to LoGoNet~\cite{logonet}, and `V.C.' refers to VirConv-S~\cite{virconv}.
}
\label{tb:kittic-car-moderate}
\renewcommand\arraystretch{1.0}
\newcommand{\tabincell}[2]{\begin{tabular}{@{}#1@{}}#2\end{tabular}}
% \scalebox{1.1}{
\setlength{\tabcolsep}{0.98mm}{
\begin{tabular}{cc|cccc|cc|ccc|ccc}
\toprule
\multicolumn{2}{c|}{\multirow{3}{*}{\textbf{Corruptions}}}  & \multicolumn{4}{c|}{\textbf{LiDAR-Only}} & \multicolumn{2}{c|}{\textbf{Camera-Only}} & \multicolumn{6}{c}{\textbf{LC Fusion}}\\
&
& 
\multirow{2}{*}{SECOND $^{\dagger}$}  & 
\multirow{2}{*}{\tabincell{c}{PointPillars $^{\dagger}$}}  &
\multirow{2}{*}{\tabincell{c}{PointRCNN $^{\dagger}$}} & 
\multirow{2}{*}{PV-RCNN $^{\dagger}$} & 
\multirow{2}{*}{SMOKE $^{\dagger}$} & 
\multirow{2}{*}{\tabincell{c}{ImVoxelNet $^{\dagger}$}} & 
\multirow{2}{*}{EPNet $^{\dagger}$} & 
\multirow{2}{*}{\tabincell{c}{Focals Conv $^{\dagger}$}} & 
\multirow{2}{*}{LoGoNet *} & 
% \multirow{2}{*}{\tabincell{c}{VirConv-S *}} & 
\multicolumn{3}{c}{RoboFusion \textbf{(Ours)}}\\
& & & & & & & & & & &  L & B & T \\
% \bottomrule
% \toprule
\midrule
\multicolumn{2}{c|}{\textbf{None}($\text{AP}_{\text{clean}}$)}  & 81.59 & 78.41 & 80.57 & 84.39 & 7.09 & 11.49 & 82.72 & 85.88 & 85.04  & \textbf{88.04}  & 87.87 & 87.60 \\
\midrule
\multicolumn{1}{c|}{}                          & Snow           & 52.34 & 36.47 & 50.36 & 52.35 & 2.47 & 0.22  & 34.58 & 34.77 & 51.45 & \textbf{85.29}  & 84.70 & 84.60 \\
\multicolumn{1}{c|}{}                          & Rain           & 52.55 & 36.18 & 51.27 & 51.58 & 3.94 & 1.24  & 36.27 & 41.30 & 55.80 &  \textbf{86.48} & 85.54 & 84.79 \\
\multicolumn{1}{c|}{}                          & Fog            & 74.10 & 64.28 & 72.14 & 79.47 & 5.63 & 1.34  & 44.35 & 44.55 & 67.53  & \textbf{85.53} & 84.00 & 84.17 \\
\multicolumn{1}{c|}{\multirow{-4}{*}{Weather}} & S.L.      & 78.32 & 62.28 & 62.78 & 79.91 & 6.00 & 10.08 & 69.65 & 80.97 & 75.54 &  \textbf{85.50} & 85.15 & 84.75 \\
\midrule
\multicolumn{1}{c|}{}                          & Density        & 80.18 & 76.49 & 80.35 & 82.79 & -    & -     & 82.09 & 84.95 & 83.68 &  \textbf{85.71} &84.34  &84.11       \\
\multicolumn{1}{c|}{}                          & Cutout         & 73.59 & 70.28 & 73.94 & 76.09 & -    & -     & 76.10 & 78.06 & 77.17 &  \textbf{83.17} &81.30  &81.21      \\
\multicolumn{1}{c|}{}                          & Crosstalk      & 80.24 & 70.85 & 71.53 & 82.34 & -    & -     & 82.10 & \textbf{85.82} & 82.00  & 84.12 &82.45  &83.07       \\
\multicolumn{1}{c|}{}                          & Gaussian (L)    & 64.90 & 74.68 & 61.20 & 65.11 & -     & -      & 60.88 & \textbf{82.14} & 61.85  & 76.56 &78.32       &76.52       \\ %Gaussian(L)
\multicolumn{1}{c|}{}                          & Uniform (L)    & 79.18 & 77.31 & 76.39 & 81.16 &  -    &  -     & 79.24 & \textbf{85.81} & 82.94  & 85.05 &83.04       &84.11       \\ %Uniform(L) 
\multicolumn{1}{c|}{}                          & Impulse (L)    & 81.43 & 78.17 & 79.78 & 82.81 &  -    &  -     & 81.63 & 85.01 & 84.66 &  85.26 & 85.06      &\textbf{85.46}       \\%Impulse (L)
\multicolumn{1}{c|}{}                          & Gaussian (C)    & -     & -     & -     & -     & 1.56 & 2.43  & 80.64 & 80.97 & 84.29 &  82.16 & \textbf{84.63}      &82.17       \\%Gaussian(C)
\multicolumn{1}{c|}{}                          & Uniform (C)     & -     & -     &   -   & -     & 2.67 & 4.85  & 81.61 & 83.38 & 84.45 & 83.30 & \textbf{85.20}      & 83.30      \\%Uniform(C)
\multicolumn{1}{c|}{\multirow{-9}{*}{Sensor}}  & Impulse (C)    & -     & -     & - & -     & 1.83 & 2.13  & 81.18 & 80.83 & 84.20 &  83.51 &  \textbf{84.55}     & 82.91      \\%Impulse (C)
\midrule
\multicolumn{1}{c|}{}                          & Moving Obj.    & 52.69 & 50.15 & 50.54 & 54.60 & 1.67 & 5.93  & \textbf{55.78} & 49.14 & 14.44 &  49.30 &  49.12     &49.90       \\%Moving Obj.
\multicolumn{1}{c|}{\multirow{-2}{*}{Motion}}  & Motion Blur    & -     & -     & - & -     & 3.51 & 4.19  & 74.71 & 81.08 & 84.52 &  84.17 & \textbf{84.56} &84.18       \\%Motion Blur
\midrule
\multicolumn{1}{c|}{}                          & Local Density  & 75.10 & 69.56 & 74.24 & 77.63 & -    & -     & 76.73 & 80.84 & 78.63 &  83.21 & 82.53 & \textbf{83.22}       \\%Local Density
\multicolumn{1}{c|}{}                          & Local Cutout   & 68.29 & 61.80 & 67.94 & 72.29 & -    & -     & 69.92 & 76.64 & 64.88 & \textbf{77.22} & 75.27 &76.23       \\%Local Cutout
\multicolumn{1}{c|}{}                          & Local Gaussian & 72.31 & 76.58 & 69.82 & 70.44 & -    & -     & 75.76 & \textbf{82.02} & 55.66 & 79.02 & 78.32 &78.33       \\%Local Gaussian
\multicolumn{1}{c|}{}                          & Local Uniform  & 80.17 & 78.04 & 77.67 & 82.09 & -    & -     & 81.71 & 84.69 & 79.94 &  \textbf{84.69} & 83.70 &84.37       \\%Local Uniform
\multicolumn{1}{c|}{\multirow{-5}{*}{Object}}  & Local Impulse  & 81.56 & 78.43 & 80.26 & 84.03 & -    & -     & 82.21 & \textbf{85.78} & 84.29  & 85.26 & 85.08 & 85.06      \\%Local Impulse
\midrule
% \bottomrule
% \toprule
\multicolumn{2}{c|}{$\textbf{Average} (\text{AP}_{\text{cor}})$}& 71.68 & 66.34 & 68.76 & 73.41 &  3.25 &  3.60 & 70.35 & 74.43 & 71.89 & \textbf{81.72} & 81.31 & 81.12 \\
% \midrule
\rowcolor[HTML]{e0ffff}
\multicolumn{2}{c|}{RCE (\%) $\downarrow$ }& 12.14 & 15.38 & 14.65 & 13.00 &  54.11 &  68.65 & 14.94 & 13.32 & 15.46 &  \textbf{7.17}  & 7.46  & 7.38  \\
\bottomrule
\end{tabular}
}
% }
\begin{tablenotes}
\footnotesize
\item[1] $^{\dagger}$: Results from Ref. \cite{Robustness3d}.
\item[2] * denotes re-implement result.
\end{tablenotes}
\end{table*}
% -----------------------------------------------------------------------------------------------------kitti-car-easy
\begin{table*}[h!t]
\scriptsize
\caption[ ]{Comparison with SOTA methods on \textbf{KITTI-C validation} set. The results are evaluated based on the \textbf{car} class with AP of $R_{40}$ at \textbf{easy} difficulty. The best one is highlighted in \textbf{bold}. `S.L.' denotes Strong Sunlight. `RCE' denotes Relative Corruption Error from Ref.\cite{Robustness3d}.}
\label{tb:kittic-car-easy}
\renewcommand\arraystretch{1.0}
\newcommand{\tabincell}[2]{\begin{tabular}{@{}#1@{}}#2\end{tabular}}
% \scalebox{0.9}{
\setlength{\tabcolsep}{0.93mm}{
\begin{tabular}{cc|cccc|cc|ccc|ccc}
\toprule
\multicolumn{2}{c|}{\multirow{3}{*}{\textbf{Corruptions}}}  & \multicolumn{4}{c|}{\textbf{Lidar-Only}} & \multicolumn{2}{c|}{\textbf{Camera-Only}} & \multicolumn{6}{c}{\textbf{LC Fusion}}\\
&
& 
\multirow{2}{*}{SECOND $^{\dagger}$}  & 
\multirow{2}{*}{\tabincell{c}{PointPillars $^{\dagger}$}}  &
\multirow{2}{*}{\tabincell{c}{PointRCNN $^{\dagger}$}} & 
\multirow{2}{*}{PV-RCNN $^{\dagger}$} & 
\multirow{2}{*}{SMOKE $^{\dagger}$} & 
\multirow{2}{*}{\tabincell{c}{ImVoxelNet $^{\dagger}$}} & 
\multirow{2}{*}{EPNet $^{\dagger}$} & 
\multirow{2}{*}{\tabincell{c}{Focals Conv $^{\dagger}$}} & 
\multirow{2}{*}{LoGoNet *} & 
% \multirow{2}{*}{\tabincell{c}{VirConv-S *}} & 
\multicolumn{3}{c}{RoboFusion \textbf{(Ours)}}\\
& & & & & & & & & & &  L & B & T \\
% \bottomrule
% \toprule
\midrule
\multicolumn{2}{c|}{\textbf{None}($\text{AP}_{\text{clean}}$)}  & 90.53 & 87.75 & 91.65 & 92.10 & 10.42 & 17.85 & 92.29 & 92.00 & 92.04 &   \textbf{93.30}    &    93.22   &   93.28    \\
\midrule
\multicolumn{1}{c|}{}                          & Snow           & 73.05 & 55.99 & 71.93 & 73.06 & 3.68  & 0.30  & 48.03 & 53.80 & 74.24 &  \textbf{88.77} & 88.18 & 88.31      \\
\multicolumn{1}{c|}{}                          & Rain           & 73.31 & 55.17 & 70.79 & 72.37 & 5.66  & 1.77  & 50.93 & 61.44 & 75.96 &  88.12      & \textbf{88.57}      & 87.75      \\
\multicolumn{1}{c|}{}                          & Fog            & 85.58 & 74.27 & 85.01 & \textbf{89.21} & 8.06  & 2.37  & 64.83 & 68.03 & 86.60 &  88.96      & 88.16      & 88.09      \\
\multicolumn{1}{c|}{\multirow{-4}{*}{Weather}} & S.L.       & 88.05 & 67.42 & 64.90 & 87.27 & 8.75  & 15.72 & 81.77 & 90.03 & 80.30 &  89.79      & 89.23      & \textbf{90.36}      \\
\midrule
\multicolumn{1}{c|}{}                          & Density        & 90.45 & 86.86 & 91.33 & 91.98 & -     & -     & 91.89 & 91.14 & 91.85 & \textbf{92.90}       &   92.08    &  92.12     \\
\multicolumn{1}{c|}{}                          & Cutout         & 81.75 & 78.90 & 83.33 & 83.40 & -     & -     & 84.17 & 83.84 & 84.20 &  \textbf{85.94}      &  85.75     &   84.75    \\
\multicolumn{1}{c|}{}                          & Crosstalk      & 89.63 & 78.51 & 77.38 & 90.52 & -     & -     & 91.30 & 92.01 & 88.15 & 91.71      &  91.54     &   \textbf{92.07}    \\
\multicolumn{1}{c|}{}                          & Gaussian (L)    & 73.21 & 86.24 & 74.28 & 74.61 & -     & -     & 66.99 & \textbf{88.56} & 64.62 &   80.96     &  84.30     &    83.23   \\
\multicolumn{1}{c|}{}                          & Uniform (L)     & 89.50 & 87.49 & 89.48 & 90.65 & -     & -     & 89.70 & 91.77 & 90.75 &   \textbf{92.89}     &   91.28    &    91.63   \\
\multicolumn{1}{c|}{}                          & Impulse (L)    & 90.70 & 87.75 & 90.80 & 91.91 & -     & -     & 91.44 & 92.10 & 91.66 &   91.90    &  91.95     &     \textbf{92.30}  \\
\multicolumn{1}{c|}{}                          & Gaussian (C)    & -     & -     & -     & -     & 2.09  & 3.74  & 91.62 & 89.51 & 91.64 &    91.94    &  \textbf{92.08}     &    91.57   \\
\multicolumn{1}{c|}{}                          & Uniform (C)     & -     & -     & -     & -     & 3.81  & 7.66  & 91.95 & 91.20 & 91.84 &   92.01    &  92.14     &    \textbf{92.93}   \\
\multicolumn{1}{c|}{\multirow{-9}{*}{Sensor}}  & Impulse (C)    & -     & -     & -     & -     & 2.57  & 3.35  & 91.68 & 89.90 & 91.65 &  91.96    &   \textbf{92.04}    &    91.33   \\
\midrule
\multicolumn{1}{c|}{}                          & Moving Obj.    & 62.64 & 58.49 & 59.29 & 63.36 & 2.69  & 9.63  & \textbf{66.32} & 54.57 & 16.83 &  53.09     &  51.94     &   51.70    \\
\multicolumn{1}{c|}{\multirow{-2}{*}{Motion}}  & Motion Blur    & -     & -     &  -    & -     & 5.39  & 6.75  & 89.65 & 91.56 & 91.96 &   91.99     &   \textbf{92.09}    &   92.06    \\
\midrule
\multicolumn{1}{c|}{}                          & Local Density  & 87.74 & 82.90 & 88.37 & 89.60 & -     & -     & 89.40 & 89.60 & 89.00 & 92.02     &  \textbf{92.42}     &    92.42   \\
\multicolumn{1}{c|}{}                          & Local Cutout   & 81.29 & 75.22 & 83.30 & 84.38 & -     & -     & 82.40 & 85.55 & 77.57 &  87.30     &  87.49     &    \textbf{87.79}   \\
\multicolumn{1}{c|}{}                          & Local Gaussian & 82.05 & 87.69 & 82.44 & 77.89 & -     & -     & 85.72 & \textbf{89.78} & 60.03 &   89.56     &  89.41     &    89.62 \\
\multicolumn{1}{c|}{}                          & Local Uniform  & 90.11 & 87.83 & 89.30 & 90.63 & -     & -     & 91.32 & 91.88 & 88.51 &   91.59    &   91.53    &     \textbf{91.75}  \\
\multicolumn{1}{c|}{\multirow{-5}{*}{Object}}  & Local Impulse  & 90.58 & 87.84 & 90.60 & 91.91 & -     & -     & 91.67 & 92.02 & 91.34 &    \textbf{92.09}    &  91.97     &     90.69  \\
\midrule
% \bottomrule
% \toprule
\multicolumn{2}{c|}{$\textbf{Average} (\text{AP}_{\text{cor}})$}& 83.10 & 77.41 & 80.78 & 83.92 & 4.74  & 5.69  & 81.63 & 83.91 & 80.93 & \textbf{88.27}      & 88.20       & 88.12      \\
% \midrule
\rowcolor[HTML]{e0ffff}
\multicolumn{2}{c|}{RCE(\%)$\downarrow$}& 8.20 & 11.78 & 11.85 & 8.87 &  54.46 &  68.07 & 11.54 & 8.78 & 12.07 &  \textbf{5.39}  & \textbf{5.39}  & 5.53  \\
\bottomrule
\end{tabular}
}
% }
\begin{tablenotes}
\footnotesize
\item[1] $^{\dagger}$: Results from Ref. \cite{Robustness3d}.
\item[2] * denotes re-implement result.

\end{tablenotes}
\end{table*}
% -----------------------------------------------------------------------------------------------------kitti-car-hard
\begin{table*}[h!t]
\scriptsize
\caption[ ]{Comparison with SOTA methods on \textbf{KITTI-C validation} set. The results are evaluated based on the \textbf{car} class with AP of $R_{40}$ at \textbf{hard} difficulty. The best one is hightlighted in \textbf{bold}. `S.L.' denotes Strong Sunlight. `RCE' denotes Relative Corruption Error from Ref.\cite{Robustness3d}.}

\label{tb:kittic-car-hard}
\renewcommand\arraystretch{1.0}
\newcommand{\tabincell}[2]{\begin{tabular}{@{}#1@{}}#2\end{tabular}}
% \scalebox{0.9}{
\setlength{\tabcolsep}{0.93mm}{
\begin{tabular}{cc|cccc|cc|ccc|ccc}
\toprule
\multicolumn{2}{c|}{\multirow{3}{*}{\textbf{Corruptions}}}  & \multicolumn{4}{c|}{\textbf{Lidar-Only}} & \multicolumn{2}{c|}{\textbf{Camera-Only}} & \multicolumn{6}{c}{\textbf{LC Fusion}}\\
&
& 
\multirow{2}{*}{SECOND $^{\dagger}$}  & 
\multirow{2}{*}{\tabincell{c}{PointPillars $^{\dagger}$}}  &
\multirow{2}{*}{\tabincell{c}{PointRCNN $^{\dagger}$}} & 
\multirow{2}{*}{PV-RCNN $^{\dagger}$} & 
\multirow{2}{*}{SMOKE $^{\dagger}$} & 
\multirow{2}{*}{\tabincell{c}{ImVoxelNet $^{\dagger}$}} & 
\multirow{2}{*}{EPNet $^{\dagger}$} & 
\multirow{2}{*}{\tabincell{c}{Focals Conv $^{\dagger}$}} & 
\multirow{2}{*}{LoGoNet *} & 
% \multirow{2}{*}{\tabincell{c}{VirConv-S *}} & 
\multicolumn{3}{c}{RoboFusion \textbf{(Ours)}}\\
& & & & & & & & & & &  L & B & T \\
% \bottomrule
% \toprule
\midrule
\multicolumn{2}{c|}{\textbf{None}($\text{AP}_{\text{clean}}$)}  & 78.57 & 75.19 & 78.06 & 82.49 & 5.57 & 9.20 & 80.16 & 83.36 & 84.31  & \textbf{85.27}   &   84.27   &83.36  \\
\midrule            
\multicolumn{1}{c|}{}                          & Snow           & 48.62 & 32.96 & 45.41 & 48.62 & 1.92 & 0.20 & 32.39 & 30.41 & 45.57 &  \textbf{64.26}      & 62.49      & 62.74      \\
\multicolumn{1}{c|}{}                          & Rain           & 48.79 & 32.65 & 45.78 & 48.20 & 3.16 & 0.99 & 34.69 & 35.71 & 50.12 & \textbf{66.07}      & 64.89      & 63.18      \\
\multicolumn{1}{c|}{}                          & Fog            & 68.93 & 58.19 & 68.05 & 75.05 & 4.56 & 1.03 & 38.12 & 39.50 & 60.47 & \textbf{80.03}      & 78.37      & 77.29      \\
\multicolumn{1}{c|}{\multirow{-4}{*}{Weather}} & S.L.       & 74.62 & 58.69 & 61.11 & 78.02 & 4.91 & 8.24 & 66.43 & 78.06 & 73.62 &  80.02      & 77.52      & \textbf{81.61}      \\
\midrule            
\multicolumn{1}{c|}{}                          & Density        & 77.04 & 72.85 & 77.58 & 81.15 & -    & -    & 79.77 & 82.38 & 81.98 &  \textbf{83.06}     & 83.03      &  83.05     \\
\multicolumn{1}{c|}{}                          & Cutout         & 70.79 & 67.32 & 71.57 & 74.60 & -    & -    & 73.95 & 76.69 & 76.18 &    76.96    & 77.00      &  \textbf{77.38}     \\
\multicolumn{1}{c|}{}                          & Crosstalk      & 76.92 & 67.51 & 69.41 & 80.98 & -    & -    & 79.54 & 83.22 & 80.36 &    82.94   & \textbf{83.22}      &  83.08     \\
\multicolumn{1}{c|}{}                          & Gaussian (L)    & 61.09 & 71.12 & 56.73 & 62.70 & -    & -    & 56.88 & \textbf{77.15} & 59.98 &   74.45    & 75.03      &  73.81     \\
\multicolumn{1}{c|}{}                          & Uniform (L)     & 75.61 & 74.09 & 72.25 & 78.93 & -    & -    & 75.92 & 81.62 & 80.68 &  81.74     &  81.79    &   \textbf{82.44}    \\
\multicolumn{1}{c|}{}                          & Impulse (L)    & 78.33 & 74.65 & 76.88 & 81.79 & -    & -    & 79.14 & \textbf{83.28} & 82.51 &    83.13    &  83.16    &    83.24   \\
\multicolumn{1}{c|}{}                          & Gaussian (C)    & -     & -     & -     & -     & 1.18 & 1.96 & 78.20 & 79.01 & 82.22 &  82.86     &   \textbf{83.05}    &    81.32   \\
\multicolumn{1}{c|}{}                          & Uniform (C)     & -     & -     & -     & -     & 2.19 & 3.90 & 79.14 & 81.39 & 82.37 & \textbf{83.22}     &  83.03     &    82.06   \\
\multicolumn{1}{c|}{\multirow{-9}{*}{Sensor}}  & Impulse (C)    & -     & -     & -     & -     & 1.52 & 1.71 & 78.51 & 78.87 & 82.16 &   82.75    &  \textbf{83.00}     &    81.59   \\
\midrule            
\multicolumn{1}{c|}{}                          & Moving Obj.    & 48.02 & 45.47 & 46.23 & 50.75 & 1.40 & 4.63 & \textbf{50.97} & 45.34 & 13.66 &   43.56    &  42.62     &    42.89   \\
\multicolumn{1}{c|}{\multirow{-2}{*}{Motion}}  & Motion Blur    & -     & -     & -     & -     & 2.95 & 3.32 & 72.49 & 77.75 & 82.50 &  \textbf{83.12}     &   83.06    &    82.92   \\
\midrule            
\multicolumn{1}{c|}{}                          & Local Density  & 71.45 & 65.70 & 71.09 & 75.39 & -    & -    & 74.36 & 77.30 & 76.83 &  \textbf{81.71}     &  81.24     &    81.15   \\
\multicolumn{1}{c|}{}                          & Local Cutout   & 63.25 & 56.69 & 63.50 & 68.58 & -    & -    & 66.53 & 72.40 & 60.62 &   71.95    &  72.07     &     \textbf{73.78}  \\
\multicolumn{1}{c|}{}                          & Local Gaussian & 68.16 & 73.11 & 65.65 & 68.03 & -    & -    & 72.71 & \textbf{78.52} & 54.02 &    76.38    &  76.41     &     76.26  \\
\multicolumn{1}{c|}{}                          & Local Uniform  & 76.67 & 74.68 & 74.37 & 80.17 & -    & -    & 78.85 & 81.99 & 77.44 &   82.04    & 82.06      &    \textbf{82.33}   \\
\multicolumn{1}{c|}{\multirow{-5}{*}{Object}}  & Local Impulse  & 78.47 & 75.18 & 77.38 & 82.33 & -    & -    & 79.79 & \textbf{83.20} & 82.21 &   82.99    & 83.16      &    82.99   \\
\midrule
% \bottomrule
% \toprule
\multicolumn{2}{c|}{$\textbf{Average} (\text{AP}_{\text{cor}})$}& 67.92 & 62.55 & 65.18 & 70.95 & 2.64 & 2.88 & 67.41 & 71.18 & 69.27 & \textbf{77.16}      & 76.81      &  76.75     \\
\rowcolor[HTML]{e0ffff}
\multicolumn{2}{c|}{RCE(\%)$\downarrow$}& 13.55 & 16.80 & 16.49 & 13.98 & 52.54 &  68.62 &  15.89 & 14.59 & 17.83 & 9.51 & 9.71  & \textbf{7.93}  \\
\bottomrule
\end{tabular}
}
% }
\begin{tablenotes}
\footnotesize
\item[1] $^{\dagger}$: Results from Ref. \cite{Robustness3d}.
\item[2] * denotes re-implement result.

\end{tablenotes}
\end{table*}
\subsubsection{More Results on the KITTI-C validation set.}
Besides the experimental results mentioned in the main text, we test our RoboFusion on KITTI-C and nuScenes-C \cite{Robustness3d} to extend our work to a wider range of noise scenarios, including Gaussian, Uniform, Impulse, Moving Object, Motion Blur, Local Density, Local Cutout, Local Gaussian, Local Uniform, and Local Impulse, as shown in Tables~\ref{tb:kittic-car-moderate}, \ref{tb:kittic-car-easy}, and \ref{tb:kittic-car-hard}. From these Tables, compared with LiDAR-only methods including SECOND \cite{SECOND}, PointPillars \cite{pointpillars}, PointRCNN \cite{pointrcnn} and PV-RCNN \cite{pvrcnn}, Camera-Only methods including Smoke \cite{liu2020smoke}, ImVoxelNet \cite{imvoxelnet}, and multi-modal methods including EPNet \cite{epnet}, Focals Conv \cite{focalconv}, and LoGoNet \cite{logonet}, our RoboFusion-L, RoboFusion-B, and RoboFusion-T consistently outperform across various noise scenarios and achieve the best overall performance. Overall, our RoboFusion demonstrates superior performance in weather-noisy (\textit{i.e.} Snow, Rain, Fog, and Strong Sunlight)  scenarios and exhibits better results across a broader range of scenarios, which shows remarkable robustness and generalizability.

\subsubsection{Performance Comparison Analysis with the LoGoNet.}
In addition, to provide a clearer analysis of performance across different noise scenarios, we present a more detailed comparative study of our RoboFusion-L and LoGoNet \cite{logonet} on the KITTI-C validation dataset, as shown in Table \ref{tab_logonet_RoboFusion}. It is worth noting that LoGoNet is a SOTA multi-modal 3D detector known for its exceptional robustness and high accuracy. \cite{Robustness3d} provides noise at varying levels, with the KITTI-C dataset including 5 severities. It is evident that our method demonstrates a high degree of robustness, exhibiting the most stable results with the variance of noise severities. %exhibiting minimal sensitivity to varying severities. 
For instance, when considering snow conditions, the performance of our RoboFusion-L shows a marginal variation from 86.69\% to 83.67\% across severities from 1 to 5. In contrast, LoGoNet's performance drops from 55.07\% to 45.02\% over the same severity range. Furthermore, in the presence of moving object noise, our method outperforms LoGoNet. In summary, our RoboFusion exhibits remarkable robustness and generalization capabilities, making it well-suited %for adapting 
to diverse noise scenarios. % and display enhanced resilience against noise interference.

\begin{table*}[h!t]
\scriptsize
    % AP_c AP_s 公式 corruptions severity
    \caption{Performance comparison of our \textcolor{red}{RoboFusion-L} with \textcolor{blue}{LoGoNet} on KITTI-C with 5 noise severities. The results are reported based on the \textbf{car} with AP of $R_{40}$ at \textbf{moderate} difficulty. `S.L.' denotes Strong Sunlight. The better one is marked in \textbf{bold}.}
    \label{tb:RoboFusion-severity}
    \centering
    \renewcommand\arraystretch{1.3}
    \newcommand{\tabincell}[2]{\begin{tabular}{@{}#1@{}}#2\end{tabular}}
    \setlength{\tabcolsep}{4.6mm}{
        \begin{tabular}{cc|ccccc|c}
        \toprule
        \multicolumn{2}{c|}{{\color[HTML]{262626} }}                              & \multicolumn{5}{c|}{Severity}         &                       \\ %\cline{3-7}
        \multicolumn{2}{c|}{\multirow{-2}{*}{{\color[HTML]{262626} Corruptions}}} & 1     & 2     & 3     & 4     & 5     & \multirow{-2}{*}{$\text{AP}_s$} \\
        \midrule
        \multicolumn{1}{c|}{}                                & Snow               & \textcolor{blue}{ 55.07} / \textcolor{red}{ \textbf{86.69}} & \textcolor{blue}{52.98} / \textcolor{red}{ \textbf{86.55}} & \textcolor{blue}{53.08} / \textcolor{red}{ \textbf{85.94}} & \textcolor{blue}{51.14} / \textcolor{red}{ \textbf{83.61}} & \textcolor{blue}{45.02} / \textcolor{red}{ \textbf{83.67}} & \textcolor{blue}{51.45} / \textcolor{red}{ \textbf{85.29}}                 \\
        \multicolumn{1}{c|}{}                                & Rain               & \textcolor{blue}{57.29} / \textcolor{red}{ \textbf{87.84}} & \textcolor{blue}{56.90} / \textcolor{red}{ \textbf{87.75}} & \textcolor{blue}{56.76} / \textcolor{red}{ \textbf{86.49}} & \textcolor{blue}{55.05} / \textcolor{red}{ \textbf{85.24}} & \textcolor{blue}{53.01} / \textcolor{red}{ \textbf{85.07}} & \textcolor{blue}{55.80} / \textcolor{red}{ \textbf{86.48}}                 \\
        \multicolumn{1}{c|}{}                                & Fog                & \textcolor{blue}{75.93} / \textcolor{red}{ \textbf{87.31}} & \textcolor{blue}{69.69} / \textcolor{red}{ \textbf{86.58}} & \textcolor{blue}{64.77} / \textcolor{red}{ \textbf{84.71}} & \textcolor{blue}{64.69} / \textcolor{red}{ \textbf{84.56}} & \textcolor{blue}{62.58} / \textcolor{red}{ \textbf{84.51}} & \textcolor{blue}{67.53} / \textcolor{red}{ \textbf{85.53}}                 \\
        \multicolumn{1}{c|}{\multirow{-4}{*}{Weather}}       & S.L.           & \textcolor{blue}{82.03} / \textcolor{red}{ \textbf{87.26}} & \textcolor{blue}{80.53} / \textcolor{red}{ \textbf{86.53}} & \textcolor{blue}{76.75} / \textcolor{red}{ \textbf{84.66}} & \textcolor{blue}{71.12} / \textcolor{red}{ \textbf{84.61}} & \textcolor{blue}{67.31} / \textcolor{red}{ \textbf{84.46}} & \textcolor{blue}{75.54} / \textcolor{red}{ \textbf{85.50}}                 \\
        \midrule
        \multicolumn{1}{c|}{}                                & Density            & \textcolor{blue}{86.60} / \textcolor{red}{ \textbf{86.81}} & \textcolor{blue}{84.59} / \textcolor{red}{ \textbf{86.59}} & \textcolor{blue}{84.05} / \textcolor{red}{ \textbf{85.60}} & \textcolor{blue}{82.74} / \textcolor{red}{ \textbf{85.27}} & \textcolor{blue}{82.42} / \textcolor{red}{ \textbf{84.30}} & \textcolor{blue}{83.68} / \textcolor{red}{ \textbf{85.71}}                 \\
        \multicolumn{1}{c|}{}                                & Cutout             & \textcolor{blue}{82.18} / \textcolor{red}{ \textbf{87.64}} & \textcolor{blue}{80.02} / \textcolor{red}{ \textbf{86.21}} & \textcolor{blue}{77.41} / \textcolor{red}{ \textbf{83.25}} & \textcolor{blue}{74.66} / \textcolor{red}{ \textbf{80.81}} & \textcolor{blue}{71.59} / \textcolor{red}{ \textbf{77.94}} & \textcolor{blue}{77.17} / \textcolor{red}{ \textbf{83.17}}                 \\
        \multicolumn{1}{c|}{}                                & Crosstalk          & \textcolor{blue}{84.22} / \textcolor{red}{ \textbf{84.41}} & \textcolor{blue}{83.38} / \textcolor{red}{ \textbf{84.38}} & \textcolor{blue}{81.41} / \textcolor{red}{ \textbf{84.13}} & \textcolor{blue}{80.78} / \textcolor{red}{ \textbf{83.79}} & \textcolor{blue}{80.22} / \textcolor{red}{ \textbf{83.90}} & \textcolor{blue}{82.00} / \textcolor{red}{ \textbf{84.12}}                 \\
        \multicolumn{1}{c|}{}                                & Gaussian (L)        & \textcolor{blue}{84.69} / \textcolor{red}{ \textbf{85.41}} & \textcolor{blue}{82.52} / \textcolor{red}{ \textbf{84.66}} & \textcolor{blue}{77.43} / \textcolor{red}{ \textbf{81.39}} & \textcolor{blue}{47.28} / \textcolor{red}{ \textbf{73.58}} & \textcolor{blue}{17.31} / \textcolor{red}{ \textbf{57.79}} & \textcolor{blue}{61.85} / \textcolor{red}{ \textbf{76.56}}                 \\
        \multicolumn{1}{c|}{}                                & Uniform (L)         & \textcolor{blue}{84.77} / \textcolor{red}{ \textbf{85.77}} & \textcolor{blue}{84.64} / \textcolor{red}{ \textbf{85.42}} & \textcolor{blue}{84.39} / \textcolor{red}{ \textbf{85.47}} & \textcolor{blue}{82.32} / \textcolor{red}{ \textbf{85.00}} & \textcolor{blue}{78.59} / \textcolor{red}{ \textbf{83.59}} & \textcolor{blue}{82.94} / \textcolor{red}{ \textbf{85.05}}                 \\
        \multicolumn{1}{c|}{}                                & Impulse (L)         & \textcolor{blue}{84.45} / \textcolor{red}{ \textbf{84.95}} & \textcolor{blue}{\textbf{84.73}} / \textcolor{red}{ 82.88} & \textcolor{blue}{\textbf{84.92}} / \textcolor{red}{ 82.20} & \textcolor{blue}{\textbf{84.63}} / \textcolor{red}{80.51} & \textcolor{blue}{\textbf{84.56}} / \textcolor{red}{80.29} & \textcolor{blue}{\textbf{84.66}} / \textcolor{red}{82.16}                 \\
        \multicolumn{1}{c|}{}                                & Gaussian (C)        & \textcolor{blue}{84.53} / \textcolor{red}{ \textbf{85.77}} & \textcolor{blue}{84.47} / \textcolor{red}{ \textbf{85.42}} & \textcolor{blue}{ 84.31} / \textcolor{red}{ \textbf{85.47}} & \textcolor{blue}{84.18} / \textcolor{red}{ \textbf{85.32}} & \textcolor{blue}{ 83.96} / \textcolor{red}{ \textbf{84.32}} & \textcolor{blue}{84.29} / \textcolor{red}{ \textbf{85.26}}                 \\
        \multicolumn{1}{c|}{}                                & Uniform (C)         & \textcolor{blue}{84.74} / \textcolor{red}{ \textbf{85.57}} & \textcolor{blue}{84.57} / \textcolor{red}{ \textbf{85.08}} & \textcolor{blue}{ \textbf{84.54}} / \textcolor{red}{ 82.96} & \textcolor{blue}{ \textbf{84.36}} / \textcolor{red}{ 82.53} & \textcolor{blue}{ \textbf{84.05}} / \textcolor{red}{ 80.36} & \textcolor{blue}{ \textbf{84.45}} / \textcolor{red}{ 83.30}                 \\
        \multicolumn{1}{c|}{\multirow{-9}{*}{Sensor}}        & Impulse (C)         & \textcolor{blue}{84.53} / \textcolor{red}{ \textbf{85.70}} & \textcolor{blue}{ \textbf{84.26}} / \textcolor{red}{ 83.63} & \textcolor{blue}{ \textbf{84.38}} / \textcolor{red}{ 83.54} & \textcolor{blue}{ \textbf{83.95}} / \textcolor{red}{ 82.42} & \textcolor{blue}{ \textbf{83.86}} / \textcolor{red}{ 82.28} & \textcolor{blue}{ \textbf{84.20}} / \textcolor{red}{ 83.51}                 \\
        \midrule
        \multicolumn{1}{c|}{}                                & Moving Obj.        & \textcolor{blue}{58.89} / \textcolor{red}{ \textbf{78.46}} & \textcolor{blue}{12.78} / \textcolor{red}{ \textbf{67.86}} & \textcolor{blue}{0.43} / \textcolor{red}{ \textbf{41.07 }} & \textcolor{blue}{0.06} / \textcolor{red}{ \textbf{36.28 }} & \textcolor{blue}{0.07} / \textcolor{red}{ \textbf{22.85 }} & \textcolor{blue}{14.44} / \textcolor{red}{ \textbf{49.30}}                 \\
        \multicolumn{1}{c|}{\multirow{-2}{*}{Motion}}        & Motion Blur        & \textcolor{blue}{84.64} / \textcolor{red}{ \textbf{85.23}} & \textcolor{blue}{84.53} / \textcolor{red}{ \textbf{84.98}} & \textcolor{blue}{84.56} / \textcolor{red}{ \textbf{84.72}} & \textcolor{blue}{ \textbf{84.45}} / \textcolor{red}{ 83.00} & \textcolor{blue}{ \textbf{84.43}} / \textcolor{red}{ 82.96} & \textcolor{blue}{ \textbf{84.52}} / \textcolor{red}{84.17}                 \\
        \midrule
        \multicolumn{1}{c|}{}                                & Local Density      & \textcolor{blue}{82.31} / \textcolor{red}{ \textbf{85.23}} & \textcolor{blue}{81.66} / \textcolor{red}{ \textbf{84.87}} & \textcolor{blue}{80.15} / \textcolor{red}{ \textbf{82.70}} & \textcolor{blue}{76.53} / \textcolor{red}{ \textbf{82.08}} & \textcolor{blue}{72.52} / \textcolor{red}{ \textbf{81.21}} & \textcolor{blue}{78.63} / \textcolor{red}{ \textbf{83.21}}                 \\
        \multicolumn{1}{c|}{}        & Local Cutout       & \textcolor{blue}{76.77} / \textcolor{red}{ \textbf{82.94}} & \textcolor{blue}{72.46} / \textcolor{red}{ \textbf{81.31}} & \textcolor{blue}{65.87} / \textcolor{red}{ \textbf{78.14}} & \textcolor{blue}{59.14} / \textcolor{red}{ \textbf{74.12}} & \textcolor{blue}{50.17} / \textcolor{red}{ \textbf{69.61}} & \textcolor{blue}{64.88} / \textcolor{red}{ \textbf{77.22}}                 \\
        \multicolumn{1}{c|}{\multirow{-2}{*}{Object}}        & Local Gaussian       & \textcolor{blue}{84.45} / \textcolor{red}{ \textbf{86.81}} & \textcolor{blue}{81.12} / \textcolor{red}{ \textbf{86.25}} & \textcolor{blue}{67.13} / \textcolor{red}{ \textbf{82.72}} & \textcolor{blue}{33.33} / \textcolor{red}{ \textbf{76.01}} & \textcolor{blue}{12.27} / \textcolor{red}{ \textbf{63.31}} & \textcolor{blue}{55.66} / \textcolor{red}{ \textbf{79.02}}                 \\
       \multicolumn{1}{c|}{}        & Local Uniform       & \textcolor{blue}{84.51} / \textcolor{red}{\textbf{85.91}} & \textcolor{blue}{84.35} / \textcolor{red}{\textbf{85.65}} & \textcolor{blue}{81.95} / \textcolor{red}{\textbf{85.23}} & \textcolor{blue}{79.62} / \textcolor{red}{\textbf{84.66}} & \textcolor{blue}{69.25} / \textcolor{red}{ \textbf{81.99}} & \textcolor{blue}{79.94} / \textcolor{red}{\textbf{84.68}}                 \\
        \multicolumn{1}{c|}{}        & Local Impulse       & \textcolor{blue}{84.53} / \textcolor{red}{ \textbf{85.65}} & \textcolor{blue}{84.47} / \textcolor{red}{ \textbf{85.13}} & \textcolor{blue}{84.32} / \textcolor{red}{ \textbf{85.18}} & \textcolor{blue}{84.40} / \textcolor{red}{ \textbf{85.16}} & \textcolor{blue}{83.72} / \textcolor{red}{ \textbf{85.16}} & \textcolor{blue}{84.29} / \textcolor{red}{ \textbf{85.25}}                 \\
        \midrule
        \multicolumn{2}{c|}{$\text{AP}_c$} & \textcolor{blue}{79.35} / \textcolor{red}{ \textbf{85.56}} & \textcolor{blue}{75.73} / \textcolor{red}{ \textbf{84.38}} & \textcolor{blue}{72.93} / \textcolor{red}{ \textbf{81.77}} & \textcolor{blue}{68.22} / \textcolor{red}{ \textbf{79.92}} & \textcolor{blue}{63.34} / \textcolor{red}{ \textbf{76.97}} & \textcolor{blue}{71.81} / \textcolor{red}{ \textbf{81.72}} \\ \midrule
                \multicolumn{2}{c|}{{\color[HTML]{262626} Clean}}                         &       &       &       &       &       & \textcolor{blue}{85.04} / \textcolor{red}{\textbf{88.04}}                 \\
        \bottomrule
        \end{tabular}
    }
    \label{tab_logonet_RoboFusion}
\end{table*}

% -----------------------------------------------------------------------------------------------------nuscenes-map
\begin{table*}[h!t]
\scriptsize
\caption[ ]{Comparison with SOTA methods on \textbf{nuScenes-C validation} set with \textbf{mAP}. `D.I.' refers to DeepInteraction~\cite{deepinteraction}. The best one is highlighted in \textbf{bold}. `S.L.' denotes Strong Sunlight. `RCE' denotes Relative Corruption Error from Ref.\cite{Robustness3d}.}
%  `P.P.' refers to PointPillars~\cite{pointpillars}, `C.P.' refers to CenterPoint~\cite{centerpoint}, `B.F.R.' refers to BEVFormer~\cite{li2022bevformer}, `T.F.' refers to TransFusion~\cite{transfusion}, `B.F.N.' refers to BEVFusion~\cite{bevfusion-mit}, . 
\label{tb:nusc-map}
\renewcommand\arraystretch{1.0}
\newcommand{\tabincell}[2]{\begin{tabular}{@{}#1@{}}#2\end{tabular}}
% \scalebox{0.9}{
\setlength{\tabcolsep}{1.2mm}{
\begin{tabular}{cc|cc|ccc|cccc|ccc}
\toprule
\multicolumn{2}{c|}{\multirow{3}{*}{\textbf{Corruptions}}}  & 
\multicolumn{2}{c|}{\textbf{Lidar-Only}} & 
\multicolumn{3}{c|}{\textbf{Camera-Only}} & 
\multicolumn{7}{c}{\textbf{LC Fusion}}\\
&
& 
\multirow{2}{*}{\tabincell{c}{PointPillars$^{\dagger}$}}  & 
% \multirow{2}{*}{SSN} &
\multirow{2}{*}{\tabincell{c}{CenterPoint$^{\dagger}$}} & 
\multirow{2}{*}{FCOS3D$^{\dagger}$} & 
% \multirow{2}{*}{PGD} & 
\multirow{2}{*}{DETR3D$^{\dagger}$} & 
\multirow{2}{*}{\tabincell{c}{BEVFormer$^{\dagger}$}} & 
\multirow{2}{*}{FUTR3D$^{\dagger}$} & 
\multirow{2}{*}{\tabincell{c}{TransFusion$^{\dagger}$}} & 
\multirow{2}{*}{BEVFusion$^{\dagger}$} & 
\multirow{2}{*}{\tabincell{c}{D.I.*}} & \multicolumn{3}{c}{RoboFusion \textbf{(Ours)}} \\
& & & & & & & & & & & L & B & T \\
% \bottomrule
% \toprule
\midrule
\multicolumn{2}{c|}{\textbf{None}($\text{AP}_{\text{clean}}$)}  & 27.69 & 59.28 & 23.86 & 34.71 & 41.65 & 64.17 & 66.38 & 68.45 & 69.90  &  \textbf{69.91}    & 69.40 & 69.09 \\
\midrule
\multicolumn{1}{c|}{}                          & Snow           & 27.57 & 55.90 & 2.01  & 5.08  & 5.73  & 52.73 & 63.30 & 62.84 & 62.36  &  \textbf{67.12}    & 66.07 & 65.96  \\
\multicolumn{1}{c|}{}                          & Rain           & 27.71 & 56.08 & 13.00 & 20.39 & 24.97 & 58.40 & 65.35 & 66.13 & 66.48   & \textbf{67.58} & 67.01 & 66.45  \\
\multicolumn{1}{c|}{}                          & Fog            & 24.49 & 43.78 & 13.53 & 27.89 & 32.76 & 53.19 & 53.67 & 54.10 & 54.79   & \textbf{67.01} & 65.54 & 64.34  \\
\multicolumn{1}{c|}{\multirow{-4}{*}{Weather}} & S.L.       & 23.71 & 54.20 & 17.20 & 34.66 & 41.68 & 57.70 & 55.14 & 64.42 & 64.93   & \textbf{67.24} & 66.71 & 66.54  \\
\midrule
\multicolumn{1}{c|}{}                          & Density        & 27.27 & 58.60 & -     & -     & -     & 63.72 & 65.77 & 67.79 &68.15        &  \textbf{69.48}    & 69.02 &68.58  \\
\multicolumn{1}{c|}{}                          & Cutout         & 24.14 & 56.28 & -     & -     & -     & 62.25 & 63.66 & 66.18 &66.23       & \textbf{69.18}     & 69.01 & 68.20 \\
\multicolumn{1}{c|}{}                          & Crosstalk      & 25.92 & 56.64 & -     & -     & -     & 62.66 & 64.67 & 67.32 &68.12        &     \textbf{68.68} & 68.04 & 68.17 \\
\multicolumn{1}{c|}{}                          & FOV lost       &  8.87 & 20.84 & -     & -     & -     & 26.32 & 24.63 & 27.17 &  \textbf{42.66}    &  39.48    &39.30  & 39.43 \\
\multicolumn{1}{c|}{}                          & Gaussian (L)    & 19.41  & 45.79 & -     & -     & -     & 58.94 & 55.10 & \textbf{60.64} & 57.46      & 57.77     &57.07  &56.00  \\
\multicolumn{1}{c|}{}                          & Uniform (L)     & 25.60 & 56.12 & -     & -     & -     & 63.21 & 64.72 & 66.81 &  \textbf{67.42}       & 64.57     &64.25  &64.99  \\
\multicolumn{1}{c|}{}                          & Impulse (L)    & 26.44 & 57.67 & -     & -     & -     & 63.43 & 65.51 & \textbf{67.54} & 67.41       & 65.64     &65.45  &65.44  \\
\multicolumn{1}{c|}{}                          & Gaussian (C)    &- &- & 3.96  & 14.86 & 15.04 & 54.96 & 64.52 & 64.44 &  66.52    &  66.73    & \textbf{66.75}  &66.53  \\
\multicolumn{1}{c|}{}                          & Uniform (C)     & -     & -     & 8.12  & 21.49 & 23.00 & 57.61 & 65.26 & 65.81 &  \textbf{65.90}    & 65.77     &65.76  &65.56  \\
\multicolumn{1}{c|}{\multirow{-9}{*}{Sensor}}  & Impulse (C)    & -     & -     & 3.55  & 14.32 & 13.99 & 55.16 & 64.37 & 64.30 &  \textbf{65.65}     &  64.82    &64.75  &64.56  \\
\midrule
\multicolumn{1}{c|}{}                          & Compensation   & 3.85  & 11.02 & -     & -     & -     & 31.87 & 9.01 & 27.57 &  39.95     &  \textbf{41.88}     &   39.54   &  41.28 \\
%\multicolumn{1}{c|}{}                          & Moving Obj.    & 19.38 & 44.30 & 10.36 & 16.63 & 20.22 & 31.87 & 9.01  & 27.57 &   -    &    -   &   -   & - \\
\multicolumn{1}{c|}{\multirow{-2}{*}{Motion}}  & Motion Blur    & -     & -     & 10.19 & 11.06 & 19.79 & 55.99 & 64.39 & 64.74 &   65.45    &    \textbf{67.21}   &   66.52   & 66.42  \\
\midrule
\multicolumn{1}{c|}{}                          & Local Density  & 26.70 & 57.55 & -     & -     & -     & 63.60 & 65.65 & 67.42 &   \textbf{67.71}    &  66.74     &   66.59   & 65.88  \\
\multicolumn{1}{c|}{}                          & Local Cutout   & 17.97 & 48.36 & -     & -     & -     & 61.85 & 63.33 & 63.41 &   65.19    &   \textbf{66.82}    &   66.53   & 66.76 \\
\multicolumn{1}{c|}{}                          & Local Gaussian & 25.93 & 51.13 & -     & -     & -     & 62.94 & 63.76 & 64.34 &   64.75    &   65.08    &   \textbf{65.17}   & 64.77 \\
\multicolumn{1}{c|}{}                          & Local Uniform  & 27.69 & 57.87 & -     & -     & -     & 64.09 & 66.20 & \textbf{67.58} &   66.44    &   66.71    &   66.19   & 65.40 \\
\multicolumn{1}{c|}{\multirow{-5}{*}{Obeject}} & Local Impulse  & 27.67 & 58.49 & -     & -     & -     & 64.02 & 66.29 & \textbf{67.91} &   67.86    &   66.53    &    66.87  & 66.67 \\
\midrule
% \bottomrule
% \toprule
\multicolumn{2}{c|}{$\textbf{Average} (\text{AP}_{\text{\text{cor}}})$} & 22.99 & 49.78 & 8.94 & 18.71 & 22.12 & 56.88 & 58.77 & 61.35 & 62.92 & \textbf{63.90} & 63.43 &  63.23  \\
\rowcolor[HTML]{e0ffff}
\multicolumn{2}{c|}{RCE (\%) $\downarrow$ }& 16.95  & 16.01  & 62.51  & 46.07  & 46.89  & 11.34  & 11.45  & 10.36  & 9.97  & 8.58  &  8.59    &  \textbf{8.47}   \\
\bottomrule
\end{tabular}
}
% }
\begin{tablenotes}
\footnotesize
\item[1] $^{\dagger}$: Results from Ref. \cite{Robustness3d}.
\item[2] * denotes re-implement result.

\end{tablenotes}
\label{tab_nuscenes_C}
\end{table*}

\subsubsection{More Results on the nuScenes-C validation set.} 
As depicted in Table \ref{tab_nuscenes_C}, compared with LiDAR-only methods including  PointPillars \cite{pointpillars}, and CenterPoint \cite{centerpoint}, Camera-Only methods FCOS3D \cite{fcos3d}, DETR3D \cite{detr3d}, and BEVFormer \cite{li2022bevformer} and multi-modal methods including FUTR3D \cite{chen2023futr3d}, TransFusion \cite{transfusion}, BEVFusion \cite{bevfusion-mit} and DeepInteraction \cite{deepinteraction}, our RoboFusion demonstrates superior performance across more noise scenarios in AD on average. For instance, our RoboFusion-L excels in 10 noise scenarios, including Weather (Snow, Rain, Fog, Strong Sunlight), Sensor (Density, Cutout, Crosstalk), Motion (Compensation, Motion Blur), and Object (Local Cutout), outperforming DeepInteraction \cite{deepinteraction} which achieves the best performance only in 5 of these noise scenarios. Overall, our method exhibits not only exceptional robustness in weather-induced noise scenarios, but also shows remarkable resilience across a broader noise include sensor, motion and object noise.

\begin{figure*}[h!t]
    \centering
    \includegraphics[width=1\linewidth]{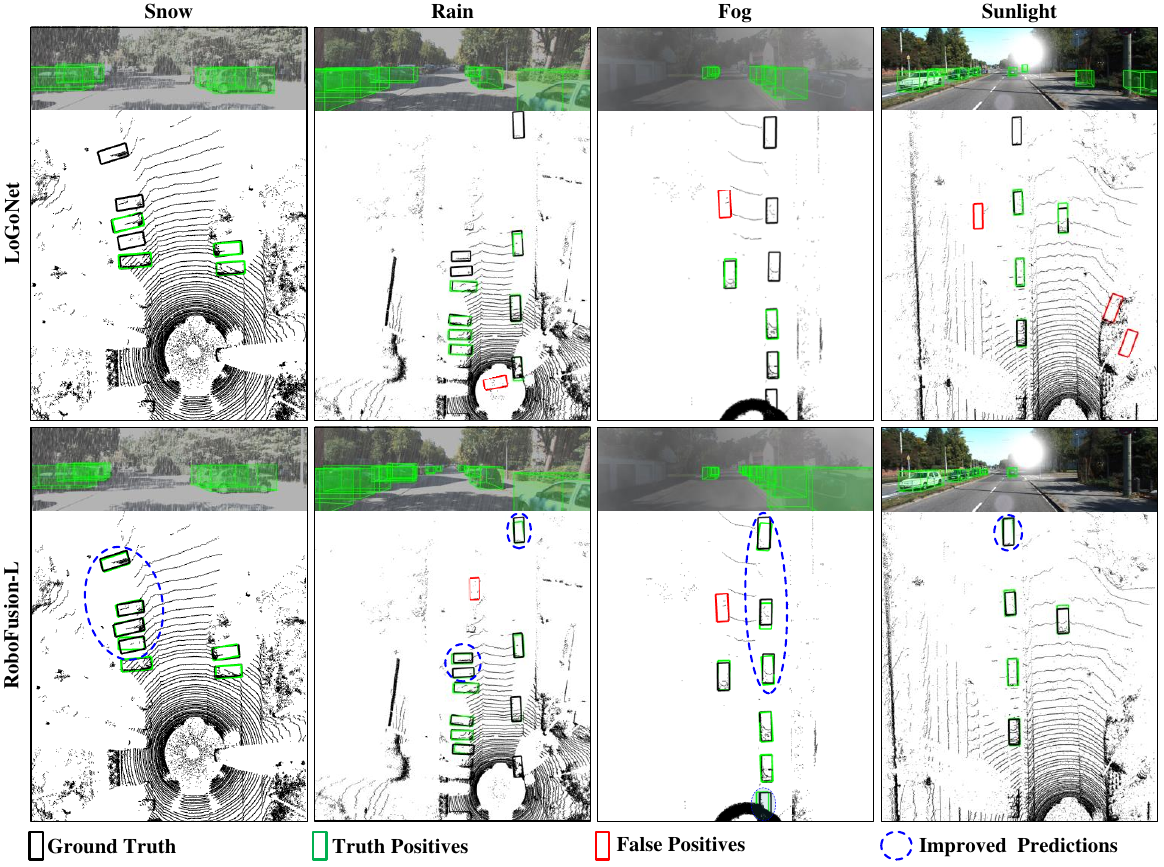}
    \caption{\textbf{Visualization Results of LoGoNet and our RoboFusion in KITTI-C dataset.} We use boxes in \textcolor{red}{red} to represent false positives, \textcolor{green}{green} boxes for truth positives, and black for the ground truth.  We use \textcolor{blue}{blue} dashed ovals to highlight the pronounced improvements in predictions.}
    \label{fig:vis-kittic}
\end{figure*}

\subsection{Visualization}
\label{sec:visualization}
As shown in Fig. \ref{fig:vis-kittic}, we provide visualization results between our RoboFusion-L and LoGoNet on the KITTI-C dataset. 
Overall, compared to SOTA methods like LoGoNet \cite{logonet}, our method enhances the robustness of multi-modal 3D object detection by leveraging the generalization capability and robustness of VFMs to mitigate OOD noisy scenarios in AD.

\subsection{More Limitations}
\label{sec:limitations}
Although we have mentioned the two main limitations in the `Conclusions' section of the main text, our RoboFusion still has other limitations. Our method does not achieve the best performance in all noisy scenarios. For instance, as shown in Table \ref{tb:kittic-car-moderate}, %compared to EPNet \cite{epnet}, 
our method does not show the best %inferior performance 
in `Moving Object' noisy scenarios.
Furthermore, we conduct experiments only on the corruption datasets \cite{Robustness3d} rather than real-world datasets. It is valuable to construct a real-world corruption dataset, but it must be an expensive work. 

%% The file named.bst is a bibliography style file for BibTeX 0.99c
\bibliographystyle{named}
\bibliography{ijcai24}

\end{document}